%% file: main.tex
\documentclass{article} %
\PassOptionsToPackage{table}{xcolor}
\usepackage{iclr2027_conference,times}

\input{math_commands.tex}

\usepackage{hyperref}
\usepackage{url}
\usepackage{amsmath}
\usepackage{amsthm}
\newtheorem{proposition}{Proposition}
\newtheorem{lemma}{Lemma}
\usepackage{graphicx, booktabs,amssymb}
\usepackage[table]{xcolor}
\usepackage{capt-of}
\usepackage{multirow}
\usepackage{siunitx}
\newcommand{\gy}{n_y}   %
\newcommand{\gx}{n_x}   %
\newcommand{\Gker}{{ G}}   %
\definecolor{ssogrow}{RGB}{229,242,255}
\definecolor{selectedrow}{RGB}{255, 224, 178}   %
\definecolor{secondaryrow}{RGB}{230, 200, 216}  %
\definecolor{tertiaryrow}{RGB}{204,229,255}
\definecolor{winmem}{RGB}{0,158,115}
\definecolor{losemem}{RGB}{213,94,0}
\definecolor{darkgreen}{RGB}{0,100,0}

\title{Beyond Quadratic Attention: Separable Gaussian Operators for ODE Vision Transformers}

\author{Carlos Boned Riera\thanks{Corresponding Author.} \\
Universitat Autònoma de Barcelona\\
Centre de Visió per Computador (CVC)\\
\texttt{cboned@cvc.uab.cat} \\
\And
David Romero Sánchez \\
Universitat Autònoma de Barcelona \\
Centre de Recerca Matemàtica (CRM) \\
\texttt{dromero@crm.cat} \\
\AND
Oriol Ramos Terrades \\
Universitat Autònoma de Barcelona\\
Centre de Visió per Computador (CVC)\\
\texttt{oriolrt@cvc.uab.cat}
}

\iclrfinalcopy %
\begin{document}

\maketitle

\begin{center}
  \includegraphics[width=0.90\linewidth]{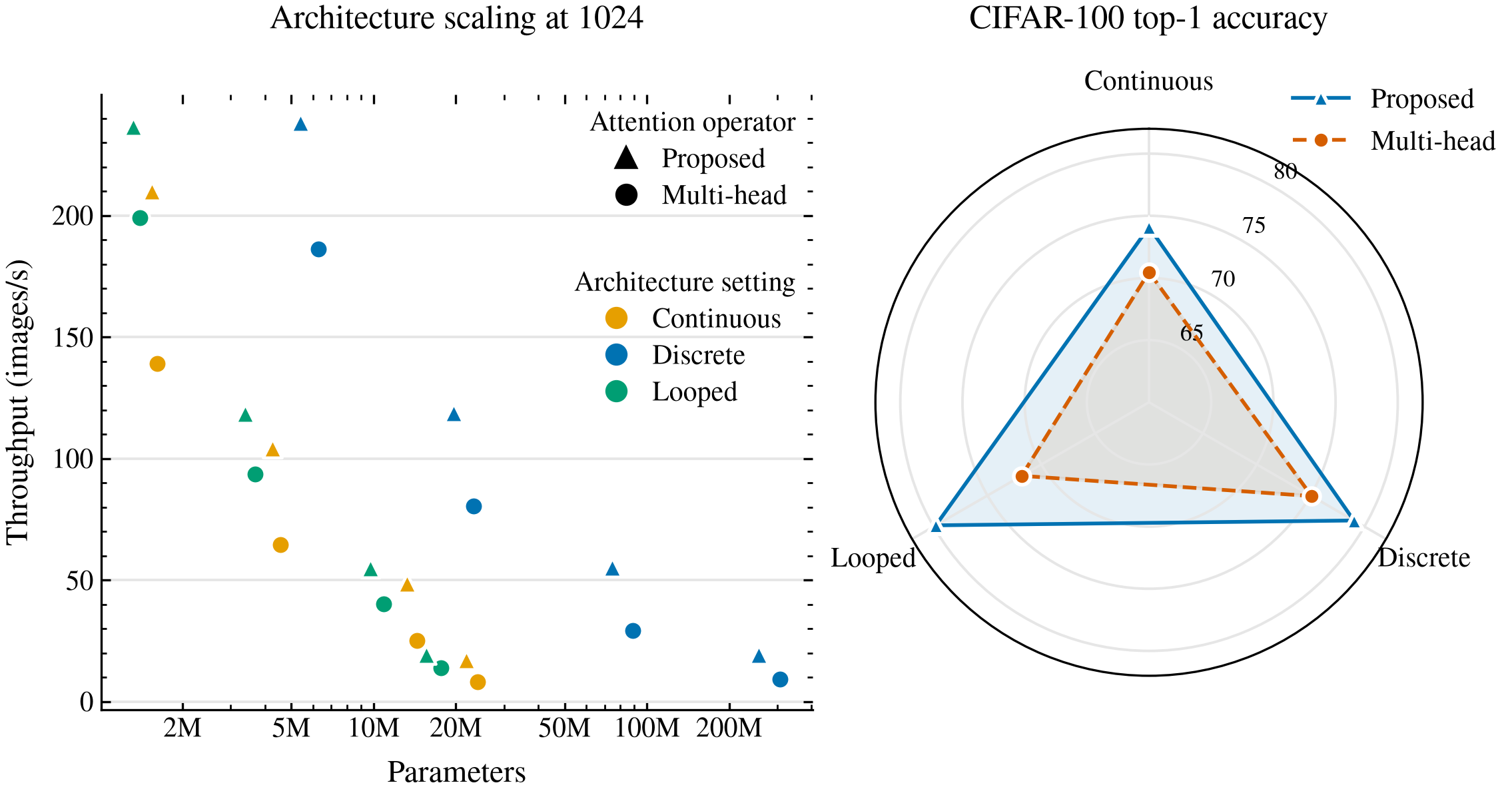}
  \captionof{figure}{{\em Left:} throughput--memory trade-off at resolution 1024 as the width of the model
increases. Points farther upward provide more throughput and points farther left allocate less memory. Our proposed method dominates multi-head attention, achieving higher throughput at lower memory. {\em Right:} comparison of the three model families with our proposed method.}
  \label{fig:efficiency-pareto}
\end{center}

\begin{abstract}
Self-attention carries a quadratic cost in token count, which limits how far Vision Transformers can scale with input resolution. We introduce SSoG, a separable sum-of-Gaussians spatial operator that replaces dense self-attention with a learned mixture of axis-factorized Gaussian kernels, reducing the per-layer cost from $\mathcal{O}(N^2)$ to subquadratic in $N$. Unlike dot-product self-attention, whose Lipschitz constant is unbounded outside a compact input domain, SSoG with fixed kernels is globally Lipschitz continuous, which, via the Picard--Lindel\"of theorem, guarantees that the continuous-depth variant has a unique solution on the whole integration interval, a global guarantee that dense-attention counterparts lack in general. On CIFAR-100 at $224\times224$, in a matched discrete-Transformer setting and in the continuous setting, SSoG improves Top-1 accuracy by 2.2--4.1 points across three model widths while using 8.8--16.6\% fewer parameters, with depth and training recipe held fixed. In a looped-Transformer setting, SSoG improves Top-1 accuracy by 5.4--8.1 points. We further probe inference-time scaling by evaluating on inputs up to $1024\times1024$ at matched batch size: in the continuous-depth (Neural ODE) architecture, SSoG reaches up to 154\% higher throughput and over 60\% lower peak memory than dense attention, with the gap widening as resolution grows, consistent with the difference in asymptotic cost. In the discrete and looped settings, SSoG also achieves over 30\% higher throughput than dense attention at this resolution. Together, these results establish SSoG as a practical, parameter-efficient, and theoretically grounded alternative to dense attention across discrete, weight-shared, and continuous-depth Transformer variants.
\end{abstract}

\section{Introduction}
\label{sec:introduction}

The self-attention mechanism, implemented through scaled dot-product attention (SDPA) \cite{AttentionAllYouNeed}, has marked a paradigm shift in AI. ViT~\cite{dosovitskiy2020image} showed that this SDPA mechanism could scale to images, and this led to remarkably impressive results across a wide range of vision tasks. Foundation models such as DINO~\cite{DINO}, SAM~\cite{SAM} for segmentation and DiT~\cite{DiT-Difussion} for diffusion, as well as VLMs and LLMs, demonstrated that this attention mechanism could scale to results beyond what seemed possible. Alongside these strengths, however, come several problems: the complexity of self-attention itself, a limited understanding of how a model learns to \emph{attend}, difficulty training these models due to theoretical issues inherent to SDPA, and the emergence of artifacts~\cite{darcet2024vision, qiu2025neurips-gated}, among others. These issues have mostly been studied in isolation. In this paper, we show that reducing the complexity of self-attention and satisfying the theoretical property required for Lipschitz continuity can be achieved by the same operator, and we study the behaviour of this operator when instantiated across discrete, looped, and continuous-depth Transformers.

The complexity of SDPA arises from the token sequence length. For $N$ tokens of dimension $d$, SDPA needs $\mathcal{O}(N^{2})$ memory and $\mathcal{O}(N^{2}d)$ computation, and in vision $N$ grows with the square of the image side, so the cost increases with the resolution. Research on context compression~\cite{zhang2024activation, liu2026sac} reduces the memory needed to store the matrix, reducing the effective sequence length, but the operation is still quadratic. State Space Models~\cite{gu2024mamba, vimamba} drop pairwise interaction altogether and are linear in $N$; however, they tend to be harder to train and imply a complete change of architecture. Looped Transformers~\cite{bViT} and Neural ODEs~\cite{zhong_neural_2022} work on a different axis: they share parameters across depth, which reduces model size but not the cost of a single evaluation. Moreover, these models are sensitive to hyperparameters, and noise in the parameter space is carried through all the optimization steps and cannot be isolated.

Beyond the cost of recurrent and continuous models, they inherit a second, more fundamental problem from dense attention itself: attention scores are computed over an unbounded domain. Since the query-key interaction is an inner product between unconstrained vectors, SDPA is not globally Lipschitz continuous~\cite{kim2021lipschitzconstantselfattention}. This is part of why the literature argues that Transformers are sensitive to hyperparameters, initialization, and learning rate~\cite{qi2023lipsformerintroducinglipschitzcontinuity}. These theoretical implications matter most for continuous-depth models. Dense attention is smooth, and hence Lipschitz on bounded sets, so the Picard--Lindel\"of theorem still guarantees a unique solution locally in time. What global Lipschitz continuity adds is existence of the solution on the whole integration interval for every initial condition, without finite-time blow-up, together with a uniform bound on how much two trajectories can drift apart when their initial conditions differ. Recurrent and continuous architectures, which apply the same block many times, are therefore the ones that benefit most from a property that dense attention lacks.

\citet{pisoni2026ssog} recently outlined SSoG, a \emph{Sum of Separable Gaussians} operator that replaces dot-product attention in standard ViTs with a small mixture of anisotropic, axis-factorized Gaussian kernels, in a \emph{fixed} and a \emph{content-steered} variant, and reported preliminary results on image classification. Because the kernel factorizes along the image axes, SSoG never builds the $N\times N$ attention matrix and is subquadratic in $N$. We study it in a regime for which it was neither designed nor analyzed: as the token mixer of looped and continuous-depth Transformers.
{ In this paper, we show that SSoG with fixed kernels is globally Lipschitz continuous by construction, precisely the property that looped and continuous-depth models require, while its content-steered variant is Lipschitz on sets of bounded token norms; we are the first to study it inside these architectures.} Unlike MHA, where tokens compete for a fixed normalization budget through content-dependent scores, fixed SSoG assigns weights from a spatial kernel of the relative offsets, so no token can take mass away from another, leaving no mechanism for an attention sink to form. We confirm this empirically on the trained models: dense attention develops a persistent sink on ImageNet, whereas SSoG does not (\appendixautorefname~\ref{sec:appendix-attention-sinks}). We benchmark this instantiation of SSoG against dense attention across discrete, looped, and continuous-depth settings, providing the first empirical and theoretical characterization of its behavior as a weight-shared, continuous-depth attention mechanism. We also find that content steering improves the discrete and looped models but not the continuous-depth one, and that the efficiency gains of SSoG grow with input resolution.

\paragraph{Contributions.}
\begin{itemize}

\item We prove that fixed SSoG is globally Lipschitz by construction, which makes continuous-depth models built on it well posed on the whole integration interval, and that the content-steered variant is Lipschitz on bounded sets (\appendixautorefname~\ref{sec:appendix-lipschitz}).

\item We use SSoG, in fixed and content-steered variants, as a drop-in replacement for dense attention in discrete, looped and continuous-depth vision Transformers, and benchmark it extensively against dense attention in each setting.

\item We study the accuracy-compute trade-off of ODE-SSoG in integration depth and number of atoms, and show its sub-quadratic operator gains over dense attention in throughput and memory as resolution grows.
\end{itemize}

\section{Method}
\label{sec:method}

We follow the formulation of~\citet{pisoni2026ssog} and restate it with the notation used in our analysis. In contrast to conventional self-attention, SSoG does not construct query-key similarities or explicitly
{ materialize an $N\times N$ attention matrix.} Instead, each head mixes value features through a small collection of learned anisotropic Gaussian spatial kernels. These kernels factorize along the $x$ and $y$ axes, so two-dimensional mixing reduces to successive one-dimensional contractions.

Each Gaussian atom, $r$, provides a compact geometric description of a spatial interaction pattern. Its center determines {\em where} information is gathered from, its spread determines {\em how local or global} the interaction is, and its mixture weight determines the contribution of that pattern to the head output.
{ Let $X\in\mathbb{R}^{B\times N\times D}$ denote the sequence of input tokens, where the $N-1$ tokens are patch tokens and one is a class token} and let $H$ be the number of heads with head dimension $D^h=D/H$. Unlike dot-product self-attention, SSoG contains no query or key projections.
{ For each head $h \in \{1,\ldots,H\}$, a learned projection $W_V^h \in \mathbb{R}^{D \times D^h}$ produces:} %
\begin{equation}
{ V^h = XW_V^h,}
\label{eq:method-values}
\end{equation}
{ Stacking all heads gives $V \in \mathbb{R}^{B \times N \times H \times D^h}$.} The class token is separated before spatial filtering, while the remaining
{ $N-1=\gy\gx$ patch tokens} are reshaped onto a $\gy\times \gx$ spatial grid.

\paragraph{Gaussian spatial kernels.}
{ Each head $h$ contains $R$ Gaussian atoms. Atom $r$ is parameterized by a center %
{ $\mu_r=(\mu_{r,x},\mu_{r,y})$, an anisotropic spread $\sigma_r=(\sigma_{r,x},\sigma_{r,y})$, and a mixture logit $\ell_r$}, from which the mixture weights $\lambda$ are obtained by softmax over atoms.}
{ For axis $a\in\{x,y\}$ and offset $\delta_a=q_a-s_a$, the unnormalized Gaussian log-kernel is:}%
\begin{equation}
{ \log \Gker^{a}(\delta_a) = -\tfrac{1}{2}\log(2\pi) - \log\sigma_a - \frac{(\delta_a-\mu_a)^2}{2\sigma_a^2}.}
\label{eq:method-gaussian-log-kernel}
\end{equation}

The kernel is normalized over source coordinates using a learned positive
temperature $\tau$,%
\begin{equation}
{
A_a(\delta) =
\operatorname{softmax}
\left(
\frac{\log\Gker^{a}(\delta)}{\tau}
\right).}
\label{eq:method-axis-kernel}
\end{equation}

{ To ensure valid, non-degenerate kernels, $\sigma_a$ and $\tau$ are obtained from unconstrained learned counterparts $\tilde{\sigma}_a$ and $\tilde{\tau}$ via a positive reparameterization, with $\sigma_{\min}>0$ bounding each axis-wise spread from below:} %
\begin{align}
{ \sigma_a}
&=
\operatorname{softplus}({ \tilde{\sigma}_a}) +\sigma_{\min}+\varepsilon \label{eq:method-constraints} \\
{ \tau}
&=
\operatorname{softplus}({ \tilde{\tau}})+\tau_0, \\
\lambda
&=
\operatorname{softmax}_{r}(\ell)
\end{align}

{ Here $\tau_0>0$ and $\varepsilon>0$ prevent over-sharpening of the axis-wise softmax and ensure numerical stability of the width constraint, respectively.}
{
Under the offset convention $\delta=q-s$, a center $\mu$ selects source positions near $s=q-\mu$, so the learned centers directly encode the relative spatial locations from which each head gathers information.

We study two variants of this operator: {\em fixed SSoG} and {\em content-steered}, described below. In {\em fixed SSoG} the geometric parameters, centers, spreads, and mixture weights, are shared across all images and query positions. In the {\em content-steered} variant, termed {\em LookAt}, bounded input-dependent perturbations adapt the geometry per query patch while preserving the separable structure.}

\subsection{Fixed SSoG}
\label{sec:method-fixed}

In fixed SSoG, the atom geometry and mixture weights are shared across images and query positions.
{ For a fixed head $h$, the value vectors are collected into $V\in\mathbb{R}^{\gy\times\gx\times D^h}$. For each atom $r$, the axis attention matrices defined entry-wise by~\eqref{eq:method-axis-kernel} are $A_{r,y}\in\mathbb{R}^{\gy\times\gy}$ and $A_{r,x}\in\mathbb{R}^{\gx\times\gx}$. The head output is:} %
{ \begin{align}
U_r
&=
A_{r,y}\,V, \\
Z
&=
\sum_{r=1}^{R}\lambda_r\,
U_r\,A_{r,x}^\top.
\label{eq:method-fixed-separable}
\end{align}}

This separable formulation reduces the asymptotic cost of spatial mixing from quadratic to subquadratic in the number of tokens:
{ for a square $n\times n$ token grid with $N-1=n^2$} and a small, fixed number of atoms $R$, applying $R$ separable Gaussian atoms through successive horizontal and vertical contractions costs $\mathcal{O}(RN^{3/2}D)$, compared with $\mathcal{O}(N^2D)$ for dense self-attention; Appendix~\ref{sec:method-complexity} gives the full derivation, including storage costs and the effect of query-dependent steering.

\subsection{Content-steered SSoG}
\label{sec:method-steered}

Fixed SSoG applies the same learned spatial geometry independently of image content. To introduce content dependence %
{ without reintroducing pairwise query-key interactions,} we additionally consider the {\em LookAt} variant. %
{ For each query position $q$, with token $x_q\in\mathbb{R}^D$ (the $q$-th row of $X$), the model predicts bounded modifications to the Gaussian centers, spreads, and mixture weights.} %
{ Three linear predictors generate the following query-dependent perturbations for each head $h$ independently:}%
\begin{align}    \label{eq:method-steering}
    \mu_{r,a}(x_q)
    &=
    \mu_{r,a}
    +
    \alpha_{\mu} M
    \tanh\bigl([W_{\mu}x_q+b_{\mu}]_{r,a}\bigr),
    \\
    \sigma_{r,a}(x_q)
    &=
    \sigma_{r,a}
    \exp\Bigl[
        \alpha_{\sigma}
        \tanh\bigl([W_{\sigma}x_q+b_{\sigma}]_{r,a}\bigr)
    \Bigr],
    \\
    \lambda_r(x_q)
    &=
    \operatorname{softmax}_{r}
    \Bigl[
        \ell_r
        +
        \alpha_{\lambda}
        \tanh\bigl([W_{\lambda}x_q+b_{\lambda}]_r\bigr)
    \Bigr].
\end{align}

{ where $W_z$ and $b_z$ are learned weights and biases, $[v]_{r,a}$ denotes the $(r,a)$-th entry of $v$ (indexed over atoms and axes), $M$ is the steering radius, so that each center moves by at most $\alpha_\mu M$ grid cells, and}%
\begin{equation}
{ \alpha_z = \operatorname{softplus}(\tilde{\alpha}_z)+\varepsilon},
\qquad z\in\{\mu,\sigma,\lambda\}.
\label{eq:method-steering-scales}
\end{equation}

The $\tanh$ nonlinearities bound the predicted modifications, while the exponential parameterization ensures that the steered spreads remain positive. Query-dependent steering does not remove the separable factorization; the computation continues to use successive one-dimensional contractions. Individual steering paths can be disabled independently for ablation.

{ The scalar steering parameters are initialized close to zero; the specific values used in each training regime are given in \S\ref{sec:results}.} The model therefore starts close to the fixed SSoG geometry and progressively activates content-dependent modulation during training. This initialization is particularly useful in the continuous-depth setting, where the operator is applied repeatedly.

\subsection{Instantiating SSoG across architectures}
\label{sec:method-instantiations}

{ SSoG is designed as a drop-in replacement for MHA: in every architecture we study, the only change relative to the corresponding baseline is replacing its attention block with SSoG, while the surrounding architecture is left unchanged. We instantiate this substitution in three settings: a standard discrete (non-recurrent) ViT, a looped/recurrent Transformer based on~\citet{bViT} that shares its attention block across depth, and a continuous-depth architecture in which SSoG defines the vector field of a Neural ODE~\cite{riera2025odevitplugplay}.}

All three settings share the same tokenization. Let an input image be $I\in\mathbb{R}^{B\times C\times H_0\times W_0}$, where $B$ is the batch size. A convolution with kernel size and stride $P$ produces a grid of $\gy \times \gx$ patch tokens, where $\gy=H_0/P$ and $\gx=W_0/P$. %
{ We denote the total sequence length by $N = \gy\gx + 1$ and the embedding dimension by $D$.}
With a class token prepended, the initial sequence is %
\begin{equation}
    X_0
    =
    {  \operatorname{PatchEmbed}(I)}
    + E_{\mathrm{pos}}
    \label{eq:method-tokenization}
\end{equation}

\paragraph{Class-token handling.}
The class token has no natural two-dimensional coordinate and is therefore excluded from Gaussian spatial filtering. Instead, it receives a lightweight global aggregation of the projected patch values $v_p \in \mathbb{R}^{D^h}$, $p = 1, \dots, N-1$: %
\begin{equation}
{ v_{\mathrm{cls}}}
=
{ \frac{1}{N-1}\sum_{p=1}^{N-1}v_p.}
\label{eq:method-cls-update}
\end{equation}

This aggregated class-token representation is concatenated with the filtered patch representations and projected through { $W_O \in \mathbb{R}^{D \times D}$}, producing the SSoG block output. Patch-to-patch mixing therefore remains entirely Gaussian, while no dense class-to-patch attention scores are introduced.
{ Here $v_{\mathrm{cls}}$ denotes the per-block value-space aggregate of the class token; in all three settings below, the class-token state accumulated across blocks is denoted $z_{\mathrm{cls}}$.}

\paragraph{Discrete ViT and looped Transformer.} In the standard discrete setting, $L$ SSoG blocks are stacked, each with its own independent parameters $\theta_1,\dots,\theta_L$. In the looped setting, a single block is used instead, with parameters $\theta$ shared across all $L$ iterations ($L$ is the depth); both follow the recurrence %
\begin{equation}
    X_{l} = X_{l-1} + \mathrm{SSoG}(X_{l-1}|\theta_l), \qquad l = 1,\dots,L,
    \label{eq:method-discrete}
\end{equation}

with $\theta_l$ independent across $l$ in the discrete case, and $\theta_l = \theta$ for all $l$ in the looped case.

\paragraph{ODE-SSoG (continuous-depth).} Rather than applying $\mathrm{SSoG}_\theta$ a fixed number of times, ODE-SSoG treats depth as continuous: SSoG serves as the token-mixing operator inside %
{ the vector field $F(\cdot,\cdot|\theta)$ of the Continuous ViT}, and the token sequence is evolved by the initial-value problem%
\begin{equation}
    \frac{dX(t)}{dt}
    =
    { F\!\left(t,X(t)|\theta\right),}
    \qquad
    X(0)=X_0,
    \qquad t\in[0,1],
    \label{eq:method-ode}
\end{equation}

integrated numerically from $t=0$ to $t=1$. As in the looped setting, the same parameters $\theta$ are reused at every function evaluation; ODE-SSoG can be seen as the continuum limit of the looped variant, where the number of iterations define the horizon in the integration path.

In all three settings, %
{ the final class-token state $z_{\mathrm{cls}}$, obtained after $L$ iterations} of~\eqref{eq:method-discrete} or by integrating ~\eqref{eq:method-ode} to $t=1$, is normalized and linearly projected to the class logits: %
\begin{equation}
    \widehat{y}
    =
    W_{\mathrm{cls}}
    { \operatorname{LN}\!\left(z_{\mathrm{cls}}\right)}
    + b_{\mathrm{cls}}.
    \label{eq:method-classifier}
\end{equation}

\section{Experiments and Results}
\label{sec:results}

In this section, we study different questions, each addressed in its own subsection. The \appendixautorefname~\ref{sec:appendix-ablations} reports further supporting ablations that isolate individual design choices without bearing directly on these questions. Unless otherwise stated, classification ablations are conducted on CIFAR-100 for 500 epochs, following the same training recipe as~\citet{DiT} and patch size 16. For the continuous models, explicit Euler integration is used; NFE denotes the number of vector field evaluations.

\begin{figure}[t]
    \centering
    \includegraphics[width=\linewidth]{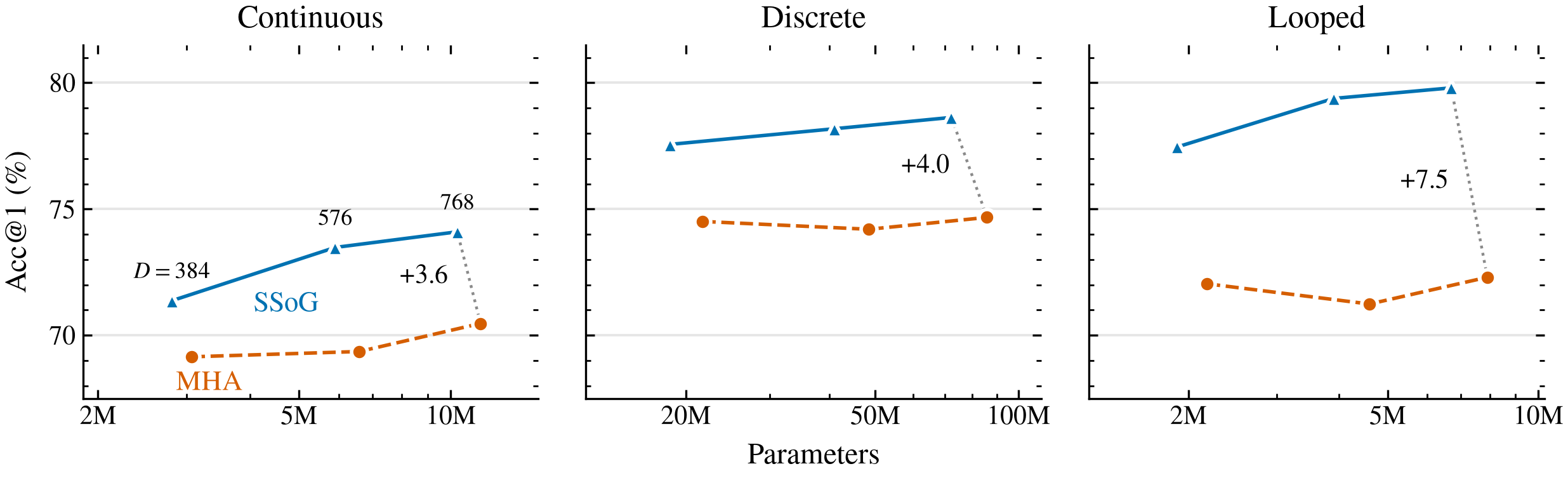}
    \caption{Width scaling on CIFAR-100, dense attention (MHA) versus SSoG, across the three architecture settings. The dotted connector and label give the Acc@1 gain of SSoG at  $D=768$.}
  \label{fig:ssog-vs-dense-width}
\end{figure}

\begin{figure}[t]
    \centering
    \includegraphics[width=\linewidth]{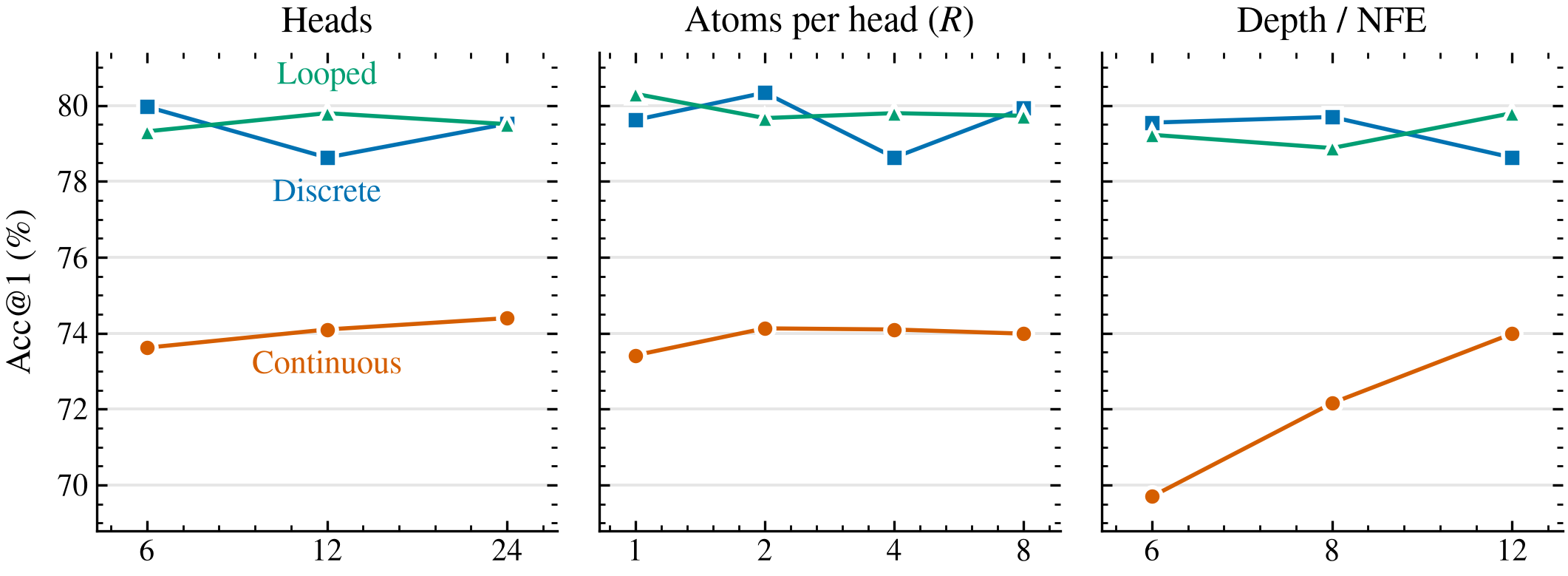}
    \caption{Ablations of SSoG on CIFAR-100 in the three architecture settings.}
  \label{fig:horizontal-ablations}
\end{figure}

\subsection{Is SSoG sensitive to different hyperparameters?}
\label{sec:results-capacity}

\figureautorefname~\ref{fig:ssog-vs-dense-width} summarizes the main comparison among models and capacity levels. In all cases the addition of the SSoG improves the results in all the families of models and in the capacity of each model within the family by at least $2.22$ Top-1 points (full results in \tableautorefname~\ref{tab:ssog-vs-dense-width}).

We report ablations over head count, depth, and atom count. \figureautorefname~\ref{fig:horizontal-ablations} shows the behavior of SSoG under different hyperparameters, with the underlying results reported in \appendixautorefname~\ref{sec:appendix-ablations}~\tableautorefname~\ref{tab:ssog-capacity}. The table shows that, at a fixed embedding dimension of 768 and four atoms per head, increasing the number of heads from 6 to 24 improves top-1 accuracy slightly in the Continuous model while there is almost no effect in the looped and discrete setting. The atom-count sweep saturates rapidly: moving from one to two atoms improves top-1 accuracy from $73.42\%$ to $74.13\%$, whereas eight atoms reach only $73.99\%$. To assess whether this saturation reflects redundancy among the learned atoms, we computed the entropy of their mixture weights: on CIFAR-100, the eight-atom model has an entropy-derived effective atom count of only $3.381$. \appendixautorefname~\ref{sec:appendix-effective-atoms} gives the formulation of the entropy used to compute the effective atoms, and \appendixautorefname~\ref{sec:appendix-sigma-floor} additionally reports the effect of the sigma floor $\sigma_{\min}$, the remaining capacity constraint on the Gaussian atoms. Two to four atoms per head therefore appear sufficient, while additional atoms are redundant.

We ran the same ablation without SSoG, using dense attention instead; the results are reported in \appendixautorefname~\ref{sec:appendix-ablations}~\tableautorefname~\ref{tab:mha-capacity}, where the variation across hyperparameters is larger than for SSoG.

\subsection{Steering: What Should Be Steered?}
\label{sec:results-steering}
We isolated query-dependent changes to the atom center $\mu$, width $\sigma$, and mixture weight $\lambda$ in \tableautorefname~\ref{tab:ssog-steering}. We used full cold steering initialization with 12 NFE as baseline. More runs can be found in the \appendixautorefname~\ref{sec:appendix-ablations}~\tableautorefname~\ref{tab:steering}. We focus on the ODE as it yielded the worst results.  Width-only steering matches the fixed model in accuracy ($74.01\%$ versus $74.22\%$). Mixture-only steering is also competitive at $73.87\%$. In contrast, jointly steering centers and widths reduces accuracy to $71.21\%$, close to full steering at $70.89\%$.  Since center-only and width-only models reach $73.33\%$ and $74.01\%$, respectively, we conjecture that the main failure mode is their joint query-dependent geometry from the division in~\eqref{eq:method-gaussian-log-kernel}. \appendixautorefname~\ref{sec:appendix-init} reports an ablation showing that a warmer initialization or a larger initialization of the gate $\tilde\alpha_\mu$ both cost additional accuracy on top of this interaction, and \appendixautorefname~\ref{sec:appendix-offset} shows that constraining the steering radius partially, but not fully, recovers it.\appendixautorefname~\ref{sec:appendix-ablations}~\tableautorefname~\ref{tab:steering} reports results of steering for the other families, where there is an improvement over the fixed SSoG.

\begin{table}[t]
  \caption{Steering-component attribution on CIFAR-100 at NFE 12 and $R=4$, cold initialization throughout. Checkmarks denote query-dependent components.}
  \label{tab:ssog-steering}
  \centering

\begin{tabular}{lcccccccc}
    \toprule
    Variant & $\mu$ & $\sigma$ & $\lambda$ & Acc@1 & Acc@3 & Acc@5 & Loss & Time (h) \\
    \midrule
    Fixed              & - & - & - & 74.22 & 88.59 & 92.38 & 1.12 & \textbf{47.77} \\
    \addlinespace[2pt]
    Full, cold         & $\checkmark$ & $\checkmark$ & $\checkmark$ & 70.89 & 86.89 & 91.30 & 1.20 & 59.48 \\
    \midrule
    Center only        & $\checkmark$ & - & - & 73.33 & 87.79 & 91.61 & 1.15 & 58.37 \\
    Width only         & - & $\checkmark$ & - & \textbf{74.01} & \textbf{88.39} & \textbf{92.16} & \textbf{1.11} & 57.42 \\
    Mixture only       & - & - & $\checkmark$ & 73.87 & 88.20 & 91.73 & 1.13 & 57.08 \\
    Center + width     & $\checkmark$ & $\checkmark$ & - & 71.21 & 86.83 & 91.13 & 1.19 & 58.61 \\
    \bottomrule
  \end{tabular}
\end{table}
\subsection{Does SSoG outperform optimized SDPA kernels as batch size and resolution grow?}
\label{sec:results-efficiency}

\begin{figure}[t]
  \centering
  \includegraphics[width=\linewidth]{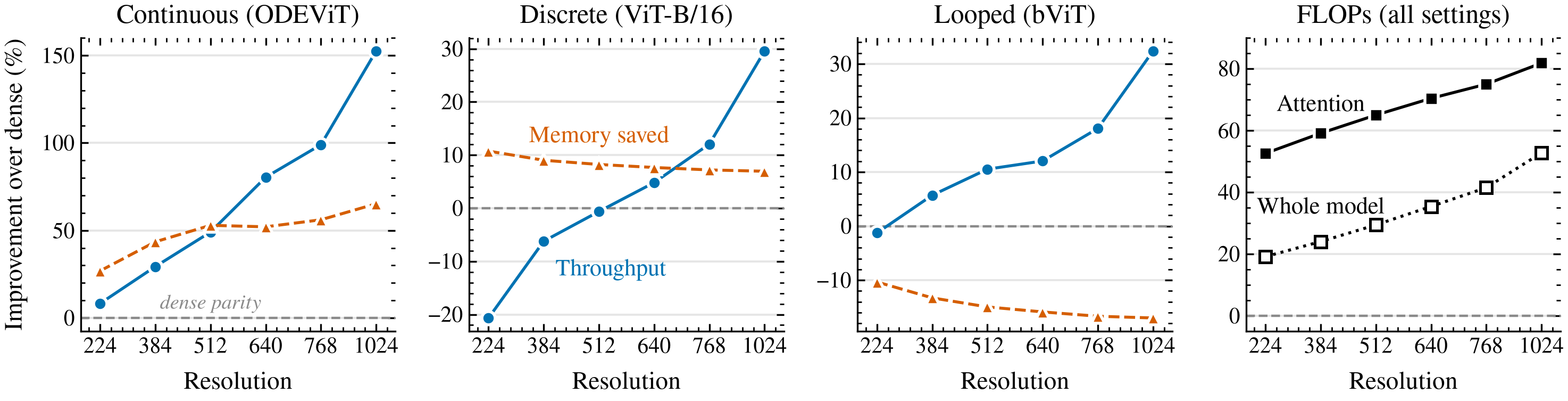}
  \caption{SSoG versus dense attention, throughput and memory improvement
  (\%) by resolution, for each of the three architecture settings
  (continuous-depth ODE-ViT, discrete ViT-B/16, looped bViT). Each point is
  the mean relative difference over batch sizes $8\to512$ at that resolution. Memory is negated (plotted as
  \% memory \emph{saved}) so both lines share one direction: above the
  dashed zero line means SSoG wins that metric at that resolution, below
  means dense wins. The rightmost panel shows the relative improvement in FLOPs for the whole model and for the attention operation alone.}
  \label{fig:delta-improvement-stress-test}
\end{figure}

We benchmark randomly initialized models in FP16 on an NVIDIA A40 using ten warm-ups and 50 measured iterations.  \figureautorefname~\ref{fig:delta-improvement-stress-test} reports a stress test of measured throughput, FLOPs and peak allocated CUDA memory at different resolutions averaged over batch sizes from 8 to 512. Full throughput, memory and FLOPs results are given in \appendixautorefname~\ref{sec:method-complexity} (\figureautorefname~\ref{fig:efficiency-dual-axis-grid-all-batches}, \tableautorefname~\ref{tab:flops}).

On the one hand, in the continuous-depth setting, the throughput of SSoG already exceeds ODE-ViT at every measured resolution, growing from $+11.0\%$ at $224\times224$ to $+154.5\%$ at $1024\times1024$. The discrete setting instead starts below parity: SSoG trails ViT-B/16 by up to $22.6\%$ at resolution 224, reaches approximate parity around resolution 512 to 640, and only clearly overtakes it from resolution 640 onward, reaching $+30.1\%$ at 1024. The looped setting sits between the two, with a positive throughput delta at every resolution, from a marginal $+2.9\%$ at 224 to $+33.1\%$ at 1024.

On the other hand, on memory, SSoG uses less memory than ODE-ViT, saving between $27\%$ and over $60\%$, and less memory than the standard ViT, saving between approximately $6\%$ and $10\%$, whereas the looped setting is the only one where SSoG costs more than dense attention. A plausible explanation is that the discrete and looped baselines use a highly optimized fused attention kernel with strong absolute memory efficiency, while the separable-Gaussian implementation used by SSoG relies on standard PyTorch ops; at high resolution, this constant-factor gap can outweigh the structural advantage of SSoG, an effect that appears earliest for the looped setting, whose dense baseline also carries less fixed parameter memory due to weight sharing. \appendixautorefname~\ref{sec:appendix-attention-scaling} and \tableautorefname~\ref{tab:appendix-attention-isolated} report the computation cost of the attention operation in isolation.

An independent sweep over Tiny, Small, Base and Large ViTs (\figureautorefname~\ref{fig:efficiency-pareto}, left) shows that SSoG is faster and lighter than MHA at every width.

\subsection{Does SSoG generalize beyond scaling?}
\label{sec:results-generalization}
Finally, \tableautorefname~\ref{tab:ssog-imagenet-accuracy} reports how the operator scales to a larger benchmark, ImageNet-1k in this case. Each model family was trained under the same experimental setup on the same hardware. SSoG matches or exceeds its dense counterpart in the continuous-depth and discrete settings, by a narrow $+0.30$-point margin for continuous-depth and a clearer $+1.42$-point margin for discrete; the looped setting is the only exception, where dense leads SSoG by $0.35$ points on top-1 accuracy. The accuracy of SSoG is competitive with dense attention across all three settings rather than specific to any one of them.~\appendixautorefname~\ref{sec:appendix-augmentation} additionally reports results on linear probing and full fine-tuning, as well as the effect of the data augmentation in the transfer setting.

\begin{table}[t]
  \caption{ImageNet-1k accuracy at $D=768$: SSoG versus dense attention (MHA).}
  \small
  \label{tab:ssog-imagenet-accuracy}
  \centering
  \begin{tabular}{lcccccc}
    \toprule
    & \multicolumn{2}{c}{Continuous} & \multicolumn{2}{c}{Discrete} & \multicolumn{2}{c}{Looped} \\
    \cmidrule(lr){2-3} \cmidrule(lr){4-5} \cmidrule(lr){6-7}
    Metric & MHA & SSoG & MHA & SSoG & MHA & SSoG \\
    \midrule
    Acc@1 & 63.04 & \textbf{64.02} & 76.47 & \textbf{77.89} & \textbf{73.72} & 73.37 \\
    Acc@3 & 79.04 & \textbf{79.62} & 89.44 & \textbf{90.59} & \textbf{88.35} & 87.93 \\
    Acc@5 & 83.40 & \textbf{84.45} & 92.31 & \textbf{93.37} & \textbf{91.90} & 91.65 \\
    \bottomrule
  \end{tabular}
\end{table}

\section{Related Work \& Discussion}
\label{sec:related-discussion}

\paragraph{Approximating attention versus replacing it.}
Most work on subquadratic attention approximates the softmax kernel instead of abandoning it, using low-rank projections, random features or kernel feature maps~\cite{wang2020linformer, choromanski2021iclr-rethinking,
katharopoulos2020icml-transformers}. What ties these together is the idea that attention acts as a kernel smoother~\cite{tsai-etal-2019-transformer}, which vision models have since adopted in several forms~\cite{han2023flatten, cai2023efficientvit}. SSoG goes the other way: it fixes the form of the kernel in advance and learns only the geometry, so the saving comes from structure and not from an approximation. Computationally the closest relatives are separable large-kernel convolutions~\cite{guo2023visual, lau2024lksa}, which split a two-dimensional depthwise kernel into cascaded one-dimensional ones for the same reason we do. What we do differently is constrain each axis kernel to a mixture of $R$ Gaussians instead of learning a free weight at every tap, and that constraint turns out to cost little, since accuracy can be saturated at few atoms per head: for instance, the continuous model with eight atoms uses an effective number of only $3.381$.
\paragraph{Gaussians as a bias, and Gaussians as the operator.}
Gaussians appear often as a spatial prior for attention. The attention maps of a trained ViT tend towards a Gaussian shape~\cite{kim2023gaussianerf}, and several models add a learned Gaussian term to the attention logits~\cite{hajimiri2026laviT}, which keeps the $N\times N$ matrix and makes the Gaussian a bias on top of attention rather than the operator itself. Closer to how we parameterize things, \citet{kundu2026gka} drop the query and key projections in favour of an RBF kernel with one bandwidth per head but still compare every pair of tokens, and concurrently \citet{huang2026gma} route tokens through Gaussian mixture components in a latent feature space for language. The operator we study is that of~\citet{pisoni2026ssog}, who proposed SSoG for standard ViTs; our contribution lies in its theoretical analysis and in its use as a weight-shared and continuous-depth token mixer. We place the Gaussians over the grid of tokens instead, which is what makes the factorization along axes possible, but it also limits how much content dependence the operator can take on. In the continuous-depth model, steering the width alone matches the fixed model ($74.01\%$ against $74.22\%$) and steering the centers alone costs $0.77$ points, while steering both drops to $71.21\%$; \eqref{eq:method-gaussian-log-kernel} suggests why: the center and the spread appear together as the ratio $(\delta_a-\mu_a)^2/2\sigma_a^2$, so when both depend on the query the numerator and the denominator can drift at the same time. We therefore recommend the fixed operator or a limited steering for this model, whereas full steering improves the discrete and looped models (\tableautorefname~\ref{tab:steering}).
\paragraph{Lipschitz continuity, weight sharing and continuous depth.}
\citet{kim2021lipschitzconstantselfattention} showed that dot-product attention is not globally Lipschitz, and the work that followed recovers the property by constraining the operator after the fact, either by substituting Lipschitz counterparts for the unstable components~\cite{qi2023lipsformerintroducinglipschitzcontinuity} or by normalizing the
layer and bounding its spectral norm~\cite{dasoulas2021lipschitznormalizationselfattentionlayers, yudin2025payattentionattentiondistribution}. In fixed SSoG it comes out of the parameterization itself, because the mixing weights depend only on bounded geometric parameters and not on inner products between unconstrained token vectors, so nothing has to be added to the loss to obtain it (Proposition~\ref{thm:fixed}); the content-steered variant retains this property only on bounded sets (Proposition~\ref{thm:steered}). This matters most when one block is applied many times, which covers both depth
recurrence, from the Universal Transformer~\cite{dehghani2018universal} and ALBERT~\cite{lan2020albert} to the fully shared single-block ViT of~\citet{bViT}, and continuous depth, from~\citet{chen2019neuralordinarydifferentialequations} to recent image classifiers~\cite{zhong_neural_2022, conf/nips/SanderAP22, riera2025odevitplugplay, belhasin2026advancing}. This second line has mostly used convolutional vector fields, which, unlike attention, are globally Lipschitz and thus satisfy the Picard--Lindel\"of theorem on the whole interval.

We hypothesize that Lipschitz continuity contributes to the accuracy gains: our three settings differ in how much the property should matter, and the results are consistent with that order. The looped model applies an unconstrained dense block repeatedly and gains the most, $7.02$ points on average and $79.80\%$ against $72.30\%$ at $D=768$. The discrete model never repeats a block and gains $3.66$ points. ODE-ViT also repeats one block, but it already relaxes the requirement to a local bound, since the JaSMin loss of~\citet{riera2025odevitplugplay} penalizes the eigenvalue distribution of the attention map and constrains the spectral radius, which may explain why its gain of $3.32$ points is closer to that of the discrete setting than to that of the looped one. The regularizer also has a computational cost. Regularizing the spectrum of the attention map requires materializing that matrix during training, which rules out a fused kernel and keeps the full $N\times N$ map in the computational graph at every function evaluation, so the memory and latency overhead grows with the NFE. What ODE-ViT obtains with a regularizer whose cost scales with the matrix it regularizes, fixed SSoG obtains from the form of the operator at no additional cost. Our efficiency benchmarks (\S\ref{sec:results-efficiency}) are measured at inference, where the regularizer is not evaluated; there, the advantage over ODE-ViT reflects the asymptotic cost of the operator together with the fact that the ODE-ViT baseline does not use a fused attention kernel (\appendixautorefname~\ref{sec:appendix-attention-scaling}).

\paragraph{Limitations.}
The margins narrow on ImageNet, where SSoG is on par with its counterparts while requiring less compute and achieving higher throughput. Part of this is expected, since a fixed spatial prior is an inductive bias and inductive biases are worth less when data is
abundant. It also bounds what the Lipschitz argument can claim, because a benefit that shrinks with dataset size behaves like regularization rather than like well-posedness. We compare only against dense attention, while linear-attention variants and state space models~\cite{gu2024mamba, vimamba} scale subquadratically as well and are the natural competitors in the regime where SSoG wins. SSoG is also $\mathcal{O}(N^{3/2})$ rather than linear (Appendix~\ref{sec:method-complexity}), so a state space model keeps an asymptotic advantage, whereas SSoG drops into an existing model without changing the rest of the architecture or the training recipe. Finally, we evaluate classification only, leaving open other tasks that could benefit from this operator.

\section{Conclusions}

We presented the first systematic study of SSoG, a separable sum-of-Gaussians alternative to dense self-attention recently outlined by~\citet{pisoni2026ssog}, as the token mixer of discrete, looped and continuous-depth Transformers. We proved that its fixed form is globally Lipschitz continuous, which guarantees a unique solution of the continuous-depth model on the whole integration interval, whereas the content-steered form is Lipschitz only on bounded sets; in both forms the cost is subquadratic in the number of tokens. On CIFAR-100, SSoG outperformed dense attention in all three settings with fewer parameters, most markedly in the looped one; on ImageNet-1k it matched or exceeded dense attention in the discrete and continuous-depth settings and trailed it slightly in the looped one. Its throughput advantage widened with resolution, and it used less memory than dense attention in the continuous-depth and discrete settings but not in the looped one, where the dense baseline benefits from a fused kernel. Future work includes a fused implementation of SSoG, tasks beyond classification, and a comparison with linear attention and state space models.

\section{Acknowledgments}
This work has been partially supported by the Spanish project PID2024-157778OB-I00, Ministerio de Ciencia e Innovación. Carlos Boned Riera is funded with the support of the FI-STEP predoctoral grants program of the Department of Research and Universities of the Generalitat of Catalonia and co-financing by the European Social Fund Plus (2025 STEP 00066)
\subsection*{AI use statement}

In this work, we used generative AI tools, namely an LLM-based coding and
writing assistant (Claude, Anthropic), to aid and polish writing, to support
research ideation and execution, and to assist with mathematical claims, as
detailed below.

\textbf{Writing.} The assistant polished the grammar and concision of
author-written paragraphs, turned results and specifications provided by the
authors into \LaTeX{} tables, figure captions and descriptive paragraphs. The authors revised all such text.

\textbf{Research ideation and execution.} The LLM wrote and debugged experiment code: training and evaluation configurations and cluster launch scripts for the ablation and transfer experiments; the throughput and peak-memory benchmark harness; the operator-level attention memory benchmark; the patch-size robustness evaluation; scripts that collect metrics from experiment logs into tables; and the plotting scripts for the figures. While auditing benchmark results, it identified a cross-model measurement interference that led us to re-run every efficiency benchmark with each model in its own process, and all reported efficiency numbers come from these isolated runs. It also proposed an analysis that we adopted: isolating the memory scaling of the attention operator from the rest of the network. The research questions, the method, the choice of experiments to report, the runs and the interpretation of results are the authors'.

\textbf{Mathematical claims.} We used AI to help validate the derivations of the Lipschitz continuity results, providing intuition and suggesting directions that supported the final lemmas and propositions. The LLM also reviewed the notation and its consistency throughout the paper.

\textbf{Verification.} We have reviewed all AI-assisted work. AI-generated code
was inspected and tested before any result was reported; for example, models re-loaded by the evaluation code were checked to reproduce their recorded validation accuracy at the training patch size, parameter counts were checked
against the saved checkpoints, and efficiency benchmarks were cross-checked by re-running them in isolation.  We take responsibility for the final content of this work, including text, claims or artifacts produced with the aid of generative AI.

\subsection*{Reproducibility statement}

We release the complete source code as anonymized supplementary material. Every experiment is specified by a modular, hierarchical Hydra configuration: a base file per dataset and architecture setting (continuous, discrete and looped; SSoG and dense-attention baselines) fixes the full training recipe, and each experiment is a short file that inherits from it and overrides only the varied quantity (width, heads, atoms, depth or NFE, or augmentation), so every reported run corresponds to exactly one versioned configuration file. Each model family has its own entry point on a shared PyTorch Lightning training pipeline, and we provide the cluster launch scripts used for every run. Training is seeded through Lightning's \texttt{seed\_everything} (including data-loader workers) and run with \texttt{deterministic=True}; the default seed is 42. Architectures and the training recipes are described in \sectionautorefname~\ref{sec:method} and \sectionautorefname~\ref{sec:results}, and the per-ablation settings in \appendixautorefname~\ref{sec:appendix-ablations}--\ref{sec:appendix-augmentation}.
Efficiency numbers come from a separate benchmark configuration and script, run on an NVIDIA A40 under \texttt{torch.inference\_mode()}, with each model benchmarked in its own process (\sectionautorefname~\ref{sec:results-efficiency} and \appendixautorefname~\ref{sec:appendix-attention-scaling}). We use only the public CIFAR-100 and ImageNet-1k datasets with their standard splits; all preprocessing and augmentation is defined in the released configurations and follows the DeiT recipe~\cite{DiT}. The table-generation and plotting scripts that produce every table and figure from the logged metrics are also included.

\bibliography{iclr2027_conference}
\bibliographystyle{iclr2027_conference}

\appendix
\section{Appendix}
\label{appendix}
\subsection{Numerical integration and computational cost}
\label{sec:method-complexity}

{All costs below are stated per sample; multiply by batch size $B$ to obtain the total cost over a mini-batch.} We integrate~\eqref{eq:method-ode} over $[0,1]$ using \texttt{torchdiffeq}, with a uniform
grid of $S$ evaluation points and the explicit Euler solver:

\begin{equation}
    X_{k+1}
    =
    X_k+\Delta t\,{ F(t_k,X_k|\theta)},
    \qquad
    \Delta t=\frac{1}{S-1}.
    \label{eq:method-euler}
\end{equation}

Euler performs $S-1$ function evaluations, so 13 grid points correspond to NFE 12.

\paragraph{Cost of one fixed SSoG evaluation.}
We count multiply-accumulate operations, reading the cost of each contraction off its index
set as the product of the ranges of all free and all summed indices. The class token adds
{ one token to the projections and is excluded from spatial filtering; since $N$ already counts it (\S\ref{sec:method-instantiations}), all costs below are written in terms of $N$.}

The value projection of~\eqref{eq:method-values} costs %
{ $\mathcal{O}(ND^{2})$}, and is the
only input projection, since SSoG forms no queries and no keys where dense attention needs
three. %
{ In the vertical contraction $U_r=A_{r,y}V$ of~\eqref{eq:method-fixed-separable}, each of the $H$ heads and $R$ atoms multiplies an $\gy\times\gy$ matrix by an $\gy\times(\gx D^h)$ matrix, giving
\begin{equation}
    H\,R\,\gy^{2}\,\gx\,D^h
    =
    \mathcal{O}\!\left(DR\,\gy^{2}\gx\right)
    \label{eq:cost-vertical}
\end{equation}
after recombining $HD^h=D$. In the horizontal contraction $\sum_r\lambda^h_rU_rA_{r,x}^\top$, the weights $\lambda^h_r$ are applied as the atom index is summed, and each head multiplies, for every atom, an $(\gy D^h)\times\gx$ matrix by an $\gx\times\gx$ matrix, giving
\begin{equation}
    H\,R\,\gy\,\gx^{2}\,D^h
    =
    \mathcal{O}\!\left(DR\,\gy\gx^{2}\right).
    \label{eq:cost-horizontal}
\end{equation}}
The class-token average of~\eqref{eq:method-cls-update} costs %
{ $\mathcal{O}(ND)$} and the
output projection %
{ $\mathcal{O}(ND^{2})$}. Adding the two contractions and using
{ $\gy\gx=N-1$, spatial mixing costs}
\begin{equation}
    { \mathcal{O}\!\left(
        DR\left(\gy^{2}\gx+\gy\gx^{2}\right)
    \right)}
    =
    { \mathcal{O}\!\left(DR\,N(\gy+\gx)\right),}
    \label{eq:method-rectangular-complexity}
\end{equation}
which for a square grid with %
{ $\gy=\gx=n$ and $N-1=n^{2}$} becomes
\begin{equation}
{ \mathcal{O}(DRn^{3})}
    =
    { \mathcal{O}\!\left(DRN^{3/2}\right).}
    \label{eq:method-square-complexity}
\end{equation}
The exponent $3/2$ rather than $1$ is the price of separability on a two-dimensional grid,
since one axis of length $\sqrt{N}$ remains inside each contraction.

Spatial mixing overtakes the projections when %
{ $DRN^{3/2}>ND^{2}$}, that is when
$R\sqrt{N}>D$, while for dense attention the token interaction overtakes them when
{ $DN^{2}>ND^{2}$}, that is when $N>D$. At $D=768$ with $R=4$ and patch size 16, dense
attention becomes interaction-bound above $N=768$ tokens, an input resolution of about 443,
whereas SSoG would only become mixing-bound beyond a resolution of 3000. Across the whole
range we benchmark, SSoG stays projection-bound while dense attention crosses into the
quadratic regime part way through, which is the mechanism behind the crossover reported in
\S\ref{sec:results-efficiency}.

\paragraph{Storage.}
Dense self-attention stores an attention matrix of size %
{ $\mathcal{O}(HN^{2})$}, and SSoG
stores none. Its axis kernels are shared across the batch and cost
$\mathcal{O}(HR(\gy^{2}+\gx^{2}))$ independently of $B$, while the atom-wise intermediates
$U$ require %
{ $\mathcal{O}(DRN)$}. The latter is linear in $N$ and explains the memory
behaviour of the looped setting in \S\ref{sec:results-efficiency}, where the dense baseline
already carries little parameter memory.

\paragraph{Cost of content-steered SSoG.}
Steering still avoids the $N\times N$ query-key matrix, but its axis kernels depend on the
batch index and on the full query position rather than being shared. Building the
query-dependent kernel tables costs %
{ $\mathcal{O}(NHR(\gy+\gx))$}, and the three linear
predictors of~\eqref{eq:method-steering} cost %
$\mathcal{O}(NDHR)$. Both are dominated by
the contraction itself, by a factor ${ D^h}$ for the tables and $\sqrt{N}/H$ for the
predictors, the latter holding at every resolution we test although only narrowly at 224.
Steering therefore has the same leading-order complexity as fixed SSoG. What changes is
arithmetic intensity: in the fixed variant each kernel entry is reused across the batch and
along one spatial axis, while under steering each entry is used once.

\subsubsection{Isolating the memory scaling of attention operation}
\label{sec:appendix-attention-scaling}

\figureautorefname~\ref{fig:delta-improvement-stress-test} shows that the memory of
 { SSoG} advantage over dense attention behaves differently across settings: it
grows with resolution in the continuous-depth setting but shrinks (discrete)
or reverses (looped) as resolution increases. Since the dense baselines of the  discrete and looped settings  both use the fused SDPA from PyTorch, while the continuous-depth baseline (ODE-ViT) does not. We isolate the attention operator itself stripped of every other confound, including patch embedding, MLP, classifier head, and weight sharing  to test whether this implementation difference alone explains the pattern.

We instantiate \texttt{SSOGAttention} (fixed, non-steered) and
\texttt{torch.nn.MultiheadAttention} (matching $D=768$, $H=12$, the exact
configuration used by the dense baselines) in isolation, feed both an
identical $[B, N, D]$ input at token counts $N$ matching each benchmarked
resolution (batch 8, patch size 16), reset the peak memory counter, and
measure only that forward call of the module.

\begin{table}[t]
  \caption{Peak allocated memory (MiB) of the attention operator alone, at
  matching token count.}
  \vspace{3mm}
  \label{tab:appendix-attention-isolated}
  \centering
  \begin{tabular}{S[table-format=4.0] S[table-format=4.0] S[table-format=3.1] S[table-format=3.1] S[table-format=1.2]}
    \toprule
    {Resolution} & {Tokens ($N$)} & {SSoG} & {Dense (fused MHA)} & {Ratio ($\times$)} \\
    \midrule
    224  & 197  & 47.6  & 29.2  & 1.63 \\
    384  & 577  & 113.0 & 56.0  & 2.02 \\
    512  & 1025 & 184.5 & 88.3  & 2.09 \\
    640  & 1601 & 281.3 & 128.1 & 2.20 \\
    768  & 2305 & 395.8 & 181.4 & 2.18 \\
    1024 & 4097 & 689.0 & 304.3 & 2.26 \\
    \bottomrule
  \end{tabular}
\end{table}

\tableautorefname~\ref{tab:appendix-attention-isolated} reports that the ratio grows monotonically with resolution: the attention
module of { SSoG}  alone uses $63\%$ more memory than the fused dense kernel at $N{=}197$, rising to
$126\%$ more at $N{=}4097$. Fitting a power law (memory $\propto N^k$) to
each series gives $k \approx 0.88$
for { SSoG} and $k \approx 0.78$ for the
fused dense kernel. We attribute this to an implementation gap rather than an inherent property of the separable formulation: the kernel construction and mixture
application of { SSoG} are built from generic PyTorch broadcast and reshape operations, whereas SDPA dispatches to a fused, memory-optimized CUDA kernel.

Furthermore, in order to isolate the computation beyond the kernels and the hardware optimization \tableautorefname~\ref{tab:flops} reports the numbers of the Flops among all settings.

\begin{table}[t]
  \caption{Forward FLOPs per image (GFLOPs, 2 per multiply-add) at dim768 with
  12 block applications. The counts are identical, to within 0.3 GFLOPs, for the continuous
  (12 ODE steps), discrete (12 blocks) and looped (12 iterations) settings.
  Attention includes the projections; element-wise operations are not counted
  for either operator. Reduction is SSoG relative to dense SDPA attention (MHA).}
  \label{tab:flops}
  \vspace{3mm}
  \centering
  \small
  \begin{tabular}{S[table-format=4.0] S[table-format=4.0] | S[table-format=4.1] S[table-format=3.1, detect-weight] S[table-format=2.1] | S[table-format=3.1] S[table-format=3.1, detect-weight] S[table-format=2.1]}
    \toprule
     & & \multicolumn{3}{c|}{Whole model} & \multicolumn{3}{c}{Attention} \\
    {Resolution} & {Tokens} & {MHA} & {SSoG} & {Reduction (\%)} & {MHA} & {SSoG} & {Reduction (\%)} \\
    \midrule
    224 & 197 & 35.1 & \bfseries 28.4 & 19.2 & 12.6 & \bfseries 6.0 & 52.6 \\
    384 & 577 & 111.0 & \bfseries 84.3 & 24.0 & 44.9 & \bfseries 18.4 & 59.1 \\
    512 & 1025 & 214.1 & \bfseries 151.1 & 29.4 & 96.8 & \bfseries 33.9 & 65.0 \\
    640 & 1601 & 368.3 & \bfseries 237.9 & 35.4 & 185.1 & \bfseries 54.9 & 70.4 \\
    768 & 2305 & 590.1 & \bfseries 345.3 & 41.5 & 326.4 & \bfseries 81.7 & 75.0 \\
    1024 & 4097 & 1319.6 & \bfseries 623.6 & 52.7 & 850.8 & \bfseries 154.9 & 81.8 \\
    \bottomrule
  \end{tabular}
\end{table}

\subsection{Attention sinks}
\label{sec:appendix-attention-sinks}

Dense attention can develop \emph{attention sinks}: tokens that receive a
disproportionate share of attention from every query, largely independent of
the input. Fixed SSoG has no mechanism to form them. Its patch-to-patch
weights depend only on the relative offset between two patches, so every key
receives the same total attention up to renormalization at the image border,
and patches never attend to the class token, which only pools. Content
steering can move a kernel, but by at most $\alpha_\mu M$ tokens, so it
cannot concentrate attention from the whole image onto one token.

We measure this on the trained models (Table~\ref{tab:attention-sinks}). A
sink requires three things at once: a key that receives much more than
uniform attention (\emph{sink score}), the same key across images
(\emph{same key}: the share of images whose most-attended key is the head's
most frequent one), and persistence across layers, which the residual
attention rollout captures because it compounds only concentration that the
heads and layers agree on.

Only the ImageNet dense model meets all three. In its worst layer the
most-attended key receives $40.6\times$ the uniform mass, it is the same token
in $91\%$ of images, and the rollout compounds it to $41.4\times$; $8.1\%$ of
patch attention also flows to the class token. The CIFAR-100 dense models each
fail at least one condition. The continuous one has strong single-head peaks
($22.9\times$ in the worst layer), but the attended key changes from image to
image (same key in $6\%$ of images) and rollout washes it out ($1.7\times$):
content-dependent focusing rather than a sink. The discrete one is
intermediate ($8.1\times$, same key in $49\%$ of images, $5.0\times$ under
rollout). The looped one has a higher same-key rate ($64\%$) but a low
worst-layer score ($3.5\times$), so consistency alone does not make a sink:
all three conditions must hold together.

Fixed SSoG stays at or below $3.3\times$ in every layer and $1.1$-$1.5\times$
under rollout in every setting. Its most-attended key is identical across
images ($100\%$) only because its content-independent kernel always peaks at
an image corner through border renormalization, a geometric effect that no
input can move. Steering raises the worst-layer score moderately
($5.3\times$ discrete, $2.9\times$ looped) and at the same time lowers the
same-key rate from $100\%$ to $60\%$ and $44\%$, while rollout stays near
uniform ($1.7\times$ and $1.6\times$). The added concentration therefore
follows the content, as in the continuous dense model, rather than forming a
fixed sink.

SSoG does not, however, avoid feature-norm artifacts altogether. On ImageNet,
$29\%$ of its final patch tokens have a norm above $3\times$ the image
median, against $3.5\%$ for the dense model. These tokens are concentrated at
the image border, which suggests a positional rather than content-driven
origin, unlike the isolated high-norm tokens associated with registers in
dense ViTs~\cite{darcet2024vision}. We therefore claim the absence of
attention sinks, not of norm artifacts.

\begin{table}[t]
  \caption{Attention-sink diagnostics of trained models on 512 validation images. First 3 settings are on CIFAR100 and the last setting on ImageNet. \emph{Sink score}: attention received by the most-attended key relative to uniform ($1$ = no sink), averaged over heads, as the mean over layers, for the worst layer, and for the residual attention rollout $\prod_l (A_l + I)/2$ of the head-averaged maps. \emph{Same key}: share of images whose most-attended key is the head's most frequent one (averaged over heads; most consistent layer); for fixed SSoG it is $100\%$ by construction, because its content-independent kernel always peaks at an image corner through border renormalization. \emph{CLS}: attention mass from patch queries to the class token (- where the model has none; SSoG patches never attend to it). \emph{High-norm}: share of final pre-norm patch tokens with norm above $3\times$ the per-image median.}
  \label{tab:attention-sinks}
  \centering
  \vspace{3mm}
  \small
  \setlength{\tabcolsep}{4pt}
  \begin{tabular}{llcccccc}
    \toprule
    & & \multicolumn{3}{c}{Sink score} & & & \\
    \cmidrule(lr){3-5}
    Setting & Operator & Mean & Worst layer & Rollout & Same key (\%) & CLS (\%) & High-norm (\%) \\
    \midrule
    \multirow{2}{*}{Continuous} & MHA & 11.3 & 22.9 & 1.7 & 6 & 0.3 & 0.0 \\
     & SSoG & 2.2 & 2.2 & 1.3 & 100 & - & 0.2 \\
    \midrule
    \multirow{3}{*}{Discrete} & MHA & 3.8 & 8.1 & 5.0 & 49 & 0.9 & 0.0 \\
     & SSoG & 2.6 & 3.3 & 1.5 & 100 & - & 0.3 \\
     & SSoG steered & 3.3 & 5.3 & 1.7 & 60 & - & 0.7 \\
    \midrule
    \multirow{3}{*}{Looped } & MHA & 2.8 & 3.5 & 2.8 & 64 & 1.0 & 0.1 \\
     & SSoG & 2.0 & 2.0 & 1.4 & 100 & - & 2.6 \\
     & SSoG steered & 2.3 & 2.9 & 1.6 & 44 & - & 0.1 \\
    \midrule
    \multirow{2}{*}{Discrete } & MHA & 20.6 & 40.6 & 41.4 & 91 & 8.1 & 3.5 \\
     & SSoG & 2.0 & 2.5 & 1.1 & 100 & - & 29.0 \\
    \bottomrule
  \end{tabular}
\end{table}

\subsection{Ablations}
\label{sec:appendix-ablations}

This appendix reports additional CIFAR-100 ablations that support the design
choices made in \S\ref{sec:results}. Unless stated otherwise,
all rows use { $D=768$}, $H=12$, $R=4$, NFE 12, and the training recipe of
\S\ref{sec:results}

\subsubsection{Sigma floor}
\label{sec:appendix-sigma-floor}

The lower bound $\sigma_{\min}$ on atom width (\eqref{eq:method-constraints})
trades off sharply local kernels against enforced smoothing. Table~\ref{tab:appendix-sigma-floor}
sweeps this floor for fixed SSoG. The default value $0.25$ performs best;
lowering it to $0.10$ costs $0.26$ points and raising it to $0.50$ costs
$0.97$ points, and the learned minimum sigma tracks the constraint in each
case, confirming that the floor is active rather than a slack bound.

\begin{table}[h]
  \caption{Effect of the sigma floor $\sigma_{\min}$ on fixed SSoG.}
  \label{tab:appendix-sigma-floor}
  \centering
  \small
  \begin{tabular}{lccc}
    \toprule
    $\sigma_{\min}$ & Acc@1 & Val.\ loss & Learned $\sigma$ min \\
    \midrule
    0.10 & 73.73 & 1.14 & 0.182 \\
    0.25 & \textbf{74.22} & \textbf{1.12} & 0.366 \\
    0.50 & 73.02 & 1.14 & 0.578 \\
    \bottomrule
  \end{tabular}
\end{table}

\begin{table}[t]
  \caption{Fixed versus steered SSoG on CIFAR-100 across the three architecture
  settings. Steps are ODE function evaluations (continuous), blocks (discrete)
  or loop iterations (looped). Time is the W\&B training runtime for 500
  epochs.}
 \label{tab:steering}
  \centering
  \small
  \vspace{5mm}
  \begin{tabular}{llcccccc}
    \toprule
    Setting & Variant & Steps & Acc@1 & Acc@3 & Acc@5 & Time (h) \\
    \midrule
    \multirow{6}{*}{Continuous}
      & Fixed   & 6  & 69.70 & 85.74 & 90.23 & 25.5 \\
      & Fixed   & 8  & 72.16 & 87.30 & 91.05 & 33.2 \\
      & Fixed   & 12 & 74.22 & 88.59 & 92.38 & 47.8 \\
    \cmidrule(l){2-7}
      & Steered & 6  & 70.36 & 86.04 & 90.67 & 31.5 \\
      & Steered & 8  & 72.00 & 87.16 & 91.42 & 40.3 \\
      & Steered & 12 & 70.89 & 86.89 & 91.30 & 59.5 \\
    \midrule
    \multirow{2}{*}{Discrete}
      & Fixed   & 12 & 78.63 & 90.12 & 93.38 & 33.40 \\
      & Steered & 12 & 81.17  & 90.21  & 91.67  & 57.81 \\
    \midrule
    \multirow{2}{*}{Looped}
      & Fixed   & 12 & 79.80 & 90.58 & 93.10 & 23.90 \\
      & Steered & 12 & 81.20  & 91.12  & 93.56  & 49.23 \\
    \bottomrule
  \end{tabular}
\end{table}

\subsubsection{Steering baselines}
\tableautorefname~\ref{tab:steering} reports the results of steered SSoG, which serve as the steering baselines. For the continuous-depth model (ODE-SSoG), steering is evaluated at different NFE; for the discrete and looped models, it is evaluated at 12 blocks and 12 iterations, respectively.

\subsubsection{Steering initialization}
\label{sec:appendix-init}

Table~\ref{tab:appendix-init} isolates how quickly the content-steered
pathway should activate. Warm initialization of full steering underperforms
cold initialization by $5.33$ points and plateaus early. Within center-only
steering,
{ moving the initialization of the gate $\tilde\alpha_\mu$ of~\eqref{eq:method-steering-scales} from $-8$ to $-4$} costs $1.80$
points despite both runs converging to a nearly identical
{ final scale $\alpha_\mu$ ($0.127$} vs.\ $0.126$), indicating that the optimization trajectory, not only
the final steering magnitude, determines the outcome.

\begin{table}[h]
  \caption{Effect of steering initialization at NFE 12, $R=4$.}
  \label{tab:appendix-init}
  \centering
  \small
  \begin{tabular}{lccc}
    \toprule
    Initialization & Final Acc@1 & Val.\ loss & Instabilities \\
    \midrule
    Cold, full steering & 70.89 & 1.2017 & None \\
    Warm, full steering & 65.56 & 1.3714 & Early plateau \\
{ Center-only, $\tilde\alpha_\mu=-8$} & \textbf{73.33} & \textbf{1.1546} & None \\
{ Center-only, $\tilde\alpha_\mu=-4$} & 71.53 & 1.2138 & None \\
    \bottomrule
  \end{tabular}
\end{table}

\subsubsection{Steering radius}
\label{sec:appendix-offset}

\tableautorefname~\ref{tab:appendix-offset} sweeps the steering radius $M$
in the full content-steered model (\eqref{eq:method-steering}). The offset
sweep is monotonic over the tested range: constraining $M$ from 7 to 4 gains
$1.59$ points, and from 4 to 2 gains a further $0.36$ points. Smaller $M$ is
{ partially compensated by a larger learned scale $\alpha_\mu$, but the effective radius $\alpha_\mu M$}
still grows with $M$ ($0.87$, $0.89$, $0.94$ for $M=2,4,7$), so the nominal
limit remains the binding constraint.

\begin{table}[h]
  \caption{Effect of the steering radius $M$, full steering, NFE 12.}
  \label{tab:appendix-offset}
  \vspace{3mm}
  \centering
  \small
  \begin{tabular}{lccc}
    \toprule
    $M$ & Acc@1 & Val.\ loss & Mean $|\mu|$ \\
    \midrule
    2 & \textbf{72.84} & \textbf{1.1385} & 0.529 \\
    4 & 72.48 & 1.1643 & 0.535 \\
    7 & 70.89 & 1.2017 & 0.475 \\
    \bottomrule
  \end{tabular}
\end{table}

\subsubsection{Discrete-setting width sweep}
\label{sec:appendix-discrete}

\tableautorefname~\ref{tab:ssog-vs-dense-width} gives the full width sweep: a 12-block
discrete transformer using SSoG at every block, compared against a matched
dense-attention transformer of identical depth, width, and training recipe.

\begin{table}[t]
  \caption{Width scaling on CIFAR-100, dense attention versus SSoG, across
  the three architecture settings. Params are per model; bold marks the
  smaller parameter count and the better Acc@1 within each row.}
  \vspace{3mm}
  \small
  \label{tab:ssog-vs-dense-width}
  \centering
  \begin{tabular}{c S[table-format=3.0] | S[table-format=2.2] S[table-format=2.1, detect-weight] | S[table-format=2.2] S[table-format=2.2, detect-weight]}
    \toprule
    {Setting} & {$D$} & {MHA Params(M)} & {SSoG Params} & {MHA Acc@1} & {SSoG Acc@1} \\
    \midrule
    \multirow{3}{*}{Continuous} & 384 & 3.07  & \bfseries 2.8  & 69.15 & \bfseries 71.37 \\
                               & 576 & 6.59  & \bfseries 5.9  & 69.36 & \bfseries 73.48 \\
                               & 768 & 11.45 & \bfseries 10.3 & 70.47 & \bfseries 74.22 \\
    \midrule
    \multirow{3}{*}{Discrete} & 384 & 21.69 & \bfseries 18.5 & 74.51 & \bfseries 77.56 \\
                              & 576 & 48.47 & \bfseries 41.0 & 74.20 & \bfseries 78.18 \\
                              & 768 & 85.86 & \bfseries 71.6 & 74.67 & \bfseries 78.63 \\
    \midrule
    \multirow{3}{*}{Looped} & 384 & 2.18 & \bfseries 1.9 & 72.04 & \bfseries 77.46 \\
                                   & 576 & 4.60 & \bfseries 3.9 & 71.24 & \bfseries 79.38 \\
                                   & 768 & 7.91 & \bfseries 6.7 & 72.30 & \bfseries 79.80 \\
    \bottomrule
  \end{tabular}
\end{table}

\begin{table*}[t]
  \caption{Capacity and depth ablations on CIFAR-100 for the three SSoG model families.
{ The Depth column gives the number of blocks for the Discrete family, the number of loop iterations for the Looped family, and the number of ODE function evaluations (NFE) for the Continuous family.} The Heads and Atoms blocks are evaluated at NFE~12 where applicable, while the Depth block varies depth or NFE at fixed $D=768$, $H=12$, $R=4$. The column highlighted in blue is the parameter varied within each block; bold marks the best metric within each block.}
  \label{tab:ssog-capacity}
  \centering
  \small
  \vspace{3mm}
  \begin{tabular}{l|c|cccc|cccc}
    \toprule
    Family & Ablation &
{ $D$} & { $H$} & $R$ & { Depth} & Par(M) & Acc@1 & Acc@3 & Acc@5 \\
    \midrule

    \multirow{11}{*}{\textsc{Discrete}}
    & \multirow{3}{*}{\textcolor{blue}{Heads}}
      & \multirow{3}{*}{768}
      & \textcolor{blue}{6}
      & \multirow{3}{*}{4}
      & \multirow{3}{*}{12}
      & \multirow{3}{*}{71.6}
      & \textbf{79.97} & 89.99 & 91.94 \\
    & & & \textcolor{blue}{12} & & & & 78.63 & \textbf{90.12} & \textbf{93.38} \\
    & & & \textcolor{blue}{24} & & & & 79.52 & 88.59 & 90.14 \\
    \cmidrule{2-10}
    & \multirow{4}{*}{\textcolor{blue}{Atoms}}
      & \multirow{4}{*}{768}
      & \multirow{4}{*}{12}
      & \textcolor{blue}{1}
      & \multirow{4}{*}{12}
      & \multirow{4}{*}{71.6}
      & 79.62 & 89.38 & 91.52 \\
    & & & & \textcolor{blue}{2} & & & \textbf{80.34} & 89.83 & 91.69  \\
    & & & & \textcolor{blue}{4} & & & 78.63& \textbf{90.12} & \textbf{93.38} \\
    & & & & \textcolor{blue}{8} & & & 79.93 & 88.67 & 90.35 \\
    \cmidrule{2-10}
    & \multirow{3}{*}{\textcolor{blue}{Depth}}
      & \multirow{3}{*}{768}
      & \multirow{3}{*}{12}
      & \multirow{3}{*}{4}
      & \textcolor{blue}{6}
      & 36.2
      & 79.55 & \textbf{90.93} & \textbf{93.89} \\
    & & & & & \textcolor{blue}{8}
      & 48.0
      & \textbf{79.70} & 90.24 & 92.66 \\
    & & & & & \textcolor{blue}{12}
      & 71.6
      & 78.63 & 90.12 & 93.38\\

    \midrule
    \multirow{11}{*}{\textsc{Looped}}
    & \multirow{3}{*}{\textcolor{blue}{Heads}}
      & \multirow{3}{*}{768}
      & \textcolor{blue}{6}
      & \multirow{3}{*}{4}
      & \multirow{3}{*}{12}
      & \multirow{3}{*}{6.7}
      & 79.32 & 90.49 & 92.85 \\
    & & & \textcolor{blue}{12} & & & & \textbf{79.80} & 90.58 & 93.10  \\
    & & & \textcolor{blue}{24} & & & & 79.51 & \textbf{90.59} & \textbf{93.32} \\
    \cmidrule{2-10}
    & \multirow{4}{*}{\textcolor{blue}{Atoms}}
      & \multirow{4}{*}{768}
      & \multirow{4}{*}{12}
      & \textcolor{blue}{1}
      & \multirow{4}{*}{12}
      & \multirow{4}{*}{6.7}
      & \textbf{80.31} & \textbf{90.68} & 93.28  \\
    & & & & \textcolor{blue}{2} & & & 79.67 & 90.57 & 93.09  \\
    & & & & \textcolor{blue}{4} & & & 79.80 & 90.58 & 93.10  \\
    & & & & \textcolor{blue}{8} & & & 79.73 & 90.60 & \textbf{93.34}  \\
    \cmidrule{2-10}
    & \multirow{4}{*}{\textcolor{blue}{Depth}}
      & \multirow{4}{*}{768}
      & \multirow{4}{*}{12}
      & \multirow{4}{*}{4}
      & \textcolor{blue}{6}
      & \multirow{4}{*}{6.7}
      & 79.23 & \textbf{91.18} & \textbf{93.96} \\
    & & & & & \textcolor{blue}{8}
      & & 78.88 & 90.18 & 93.17 \\
    & & & & & \textcolor{blue}{12}
      & & \textbf{79.80} & 90.58 & 93.10 \\

    \midrule
    \multirow{11}{*}{\textsc{Continuous}}
    & \multirow{3}{*}{\textcolor{blue}{Heads}}
      & \multirow{3}{*}{768}
      & \textcolor{blue}{6}
      & \multirow{3}{*}{4}
      & \multirow{3}{*}{12}
      & \multirow{3}{*}{10.3}
      & 73.62 & 88.39 & 92.11 \\
    & & & \textcolor{blue}{12} & & & & 74.22 & \textbf{88.59} & \textbf{92.38} \\
    & & & \textcolor{blue}{24} & & & & \textbf{74.40} & 88.44 & 92.37 \\
    \cmidrule{2-10}
    & \multirow{4}{*}{\textcolor{blue}{Atoms}}
      & \multirow{4}{*}{768}
      & \multirow{4}{*}{12}
      & \textcolor{blue}{1}
      & \multirow{4}{*}{12}
      & \multirow{4}{*}{10.3}
      & 73.42 & 88.25 & 92.05 \\
    & & & & \textcolor{blue}{2} & & & \textbf{74.13} & 88.36 & 91.95 \\
    & & & & \textcolor{blue}{4} & & & 74.22 & \textbf{88.59} & 92.38  \\
    & & & & \textcolor{blue}{8} & & & 73.99 & 88.58 & \textbf{92.47} \\
    \cmidrule{2-10}
    & \multirow{4}{*}{\textcolor{blue}{Depth}}
      & \multirow{4}{*}{768}
      & \multirow{4}{*}{12}
      & \multirow{4}{*}{4}
      & \textcolor{blue}{6}
      & \multirow{4}{*}{10.3}
      & 69.70 & 85.74 & 90.23  \\
    & & & & & \textcolor{blue}{8}
      & & 72.16 & 87.30 & 91.05  \\
    & & & & & \textcolor{blue}{12}
      & & 74.22 & 88.59 & 92.38 \\
    & & & & & \textcolor{blue}{16}
      & & \textbf{74.92} & \textbf{89.02} & \textbf{92.46} \\

    \bottomrule
  \end{tabular}
\end{table*}

\begin{table*}[t]
  \caption{MHA capacity and depth ablations on CIFAR-100 for the Discrete MHA (ViT) and Looped MHA (bViT) families. The column highlighted in blue is the parameter varied within each block; bold marks the best metric within each block.}
  \label{tab:mha-capacity}
  \centering
  \small
  \vspace{3mm}
  \begin{tabular}{l|c|ccc|cccc}
    \toprule
    Family & Ablation &
{ $D$} & { $H$} & Depth & Params & Acc@1 & Acc@3 & Acc@5 \\
    \midrule

    \multirow{6}{*}{\textsc{Discrete}}
    & \multirow{3}{*}{\textcolor{blue}{Heads}}
      & \multirow{3}{*}{768}
      & \textcolor{blue}{6}
      & \multirow{3}{*}{12}
      & \multirow{3}{*}{85.9M}
      & \textbf{75.63} & \textbf{87.51} & \textbf{90.55} \\
    & & & \textcolor{blue}{12}
      & & &
      74.67 & 86.89 & 90.19 \\
    & & & \textcolor{blue}{24}
      & & &
      73.70 & 86.0 & 89.30 \\

    \cmidrule{2-9}

    & \multirow{3}{*}{\textcolor{blue}{Depth}}
      & \multirow{3}{*}{768}
      & \multirow{3}{*}{12}
      & \textcolor{blue}{6}
      & 43.3M
      & 70.47 & 84.50 & 88.61 \\
    & & & & \textcolor{blue}{8}
      & 57.5M
      & 72.22 & 85.68 & 89.60 \\
    & & & & \textcolor{blue}{12}
      & 85.9M
      & \textbf{74.67} & \textbf{86.89} & \textbf{90.19} \\

    \midrule

    \multirow{7}{*}{\textsc{Looped}}
    & \multirow{3}{*}{\textcolor{blue}{Heads}}
      & \multirow{3}{*}{768}
      & \textcolor{blue}{6}
      & \multirow{3}{*}{12}
      & \multirow{3}{*}{7.9M}
      & \textbf{73.18} & \textbf{85.80} & \textbf{89.41} \\
    & & & \textcolor{blue}{12}
      & & &
      72.30 & 84.37 & 88.33 \\
    & & & \textcolor{blue}{24}
      & & &
      70.07 & 83.13 & 86.89 \\

    \cmidrule{2-9}

    & \multirow{3}{*}{\textcolor{blue}{Depth}}
      & \multirow{3}{*}{768}
      & \multirow{3}{*}{12}
      & \textcolor{blue}{6}
      & \multirow{3}{*}{7.9M}
      & 68.72 & 83.44 & 87.69 \\
    & & & & \textcolor{blue}{8}
      & &
      70.99 & \textbf{84.57} & \textbf{88.60} \\
    & & & & \textcolor{blue}{12}
      & &
      \textbf{72.30} & 84.37 & 88.33 \\

    \bottomrule
  \end{tabular}
\end{table*}

\subsubsection{Resolution Robustness}

In order to validate the robustness of our method whe test the trained models with different patch size. \tableautorefname~\ref{tab:patch-size-robustness} reports the results of models trained at patch 16 (224px, $14\times14$ tokens)
evaluated at patch 14 and 8 without retraining. Patch-embedding kernels are PI-resized and position embeddings bicubically interpolated for every model; the rescaled SSoG row also scales $\mu$, $\sigma$ by the grid ratio so each Gaussian covers the same image.

\begin{table}[t]
  \caption{Patch-size robustness on CIFAR-100: models trained at patch 16 (224px, $14\times14$ tokens) evaluated at patch 14 and 8 without retraining. Patch-embedding kernels are PI-resized and position embeddings bicubically interpolated for every model; the rescaled SSoG row also scales $\mu$, $\sigma$ by the grid ratio so each Gaussian covers the same image area. Acc@1 (\%), change vs.\ patch 16 in parentheses.}
  \vspace{3mm}
  \label{tab:patch-size-robustness}
  \centering
  \small
  \begin{tabular}{ll S[table-format=2.2, detect-weight] S[table-format=2.2, detect-weight]@{\,(}S[table-format=+1.2, retain-explicit-plus]@{)\hspace{2\tabcolsep}}S[table-format=2.2, detect-weight]@{\,(}S[table-format=+2.2, retain-explicit-plus]@{)\hspace{\tabcolsep}}}
    \toprule
    & & {Patch 16} & \multicolumn{2}{c}{Patch 14} & \multicolumn{2}{c}{Patch 8} \\
    {Setting} & {Operator} & {($14\times14$)} & \multicolumn{2}{c}{($16\times16$)} & \multicolumn{2}{c}{($28\times28$)} \\
    \midrule
    \multirow{3}{*}{Continuous} & MHA & 70.47 & 70.23 & +0.80 & 58.05 & -11.38 \\
      & SSoG & \bfseries 74.20 & \bfseries 74.85 & +0.65 & 57.44 & -16.76 \\
      & SSoG (rescaled $\mu,\sigma$) & \bfseries 74.20 & 74.30 & +0.10 & \bfseries 64.39 & -9.81 \\
    \midrule
    \multirow{3}{*}{Discrete} & MHA & 74.67 & 74.91 & -0.18 & 56.92 & -18.17 \\
      & SSoG & \bfseries 78.63 & 79.34 & +0.71 & 63.49 & -15.53 \\
      & SSoG (rescaled $\mu,\sigma$) & \bfseries 78.63 & \bfseries 79.48 & +0.85 & \bfseries 72.09 & -6.93 \\
    \midrule
    \multirow{3}{*}{Looped} & MHA & 72.30 & 72.21 & -0.09 & 52.24 & -20.06 \\
      & SSoG & \bfseries 79.80 & \bfseries 80.98 & +1.18 & 73.56 & -6.74 \\
      & SSoG (rescaled $\mu,\sigma$) & \bfseries 79.80 & 80.12 & +0.42 & \bfseries 73.59 & -6.71 \\
    \bottomrule
  \end{tabular}
\end{table}

\subsection{Effective atoms}
\label{sec:appendix-effective-atoms}

To measure how many of the $R$ Gaussian atoms a head actually uses, we take
its learned mixture weights ${ \lambda^h_r}=\operatorname{softmax}_r({ \ell^h_r})$
and compute their Shannon entropy and its exponential,
\begin{equation}
    { R^h_{\mathrm{eff}}}
    =
    \exp\!\Big(-\sum_{r=1}^{R}{ \lambda^h_r}\log{ \lambda^h_r}\Big),
    \qquad
    { \overline{R}_{\mathrm{eff}}}=\frac{1}{H}\sum_{h=1}^{H}{ R^h_{\mathrm{eff}}}.
    \label{eq:effective-atoms}
\end{equation}
${ R^h_{\mathrm{eff}}}$ ranges from $1$ (all mass on one atom) to $R$ (uniform
mixture), and we report its mean over heads. It is computed on the fixed
per-head weights, before content-dependent steering, so it shows how the
model spends its atom budget rather than any single query's mixture.

\subsubsection{ImageNet width confirmation}
\label{sec:appendix-imagenet}

Table~\ref{tab:appendix-imagenet} reports the ImageNet width sweep for fixed
SSoG at constant
{ head dimension $D^h=64$.} Best-epoch Top-1 improves from $55.65\%$ to
$62.43\%$ to $64.02\%$
{ as $D$ increases from 384 to 576 to 768}, with
diminishing gains from 576 to 768.

\begin{table}[h]
  \caption{ImageNet width sweep, fixed SSoG, $R=4$, NFE 12.}
  \vspace{3mm}
  \label{tab:appendix-imagenet}
  \centering
  \small
  \begin{tabular}{lccc}
    \toprule
{ $D$} & { $H$} & Acc@1 & Acc@5 \\
    \midrule
    384 & 6  & 55.65 (245)& 78.46 \\
    576 & 9  & 62.43 (273)& 83.37 \\
    768 & 12 & \textbf{64.02 (254)} & 84.45 \\
    \bottomrule
  \end{tabular}
  \vspace{1mm}
\end{table}

\subsection{Augmentation and ImageNet-to-CIFAR-100 transfer}
\label{sec:appendix-augmentation}

Table~\ref{tab:appendix-augmentation-scratch} sweeps training augmentation
strength for fixed SSoG trained from scratch on CIFAR-100: stronger
augmentation is preferred when learning the full representation from random
initialization. Table~\ref{tab:appendix-transfer} instead adapts to CIFAR-100, either through a linear probe or full fine-tuning; here weaker augmentation is preferred, reversing the from-scratch trend, consistent with adapting rather than learning a representation from random initialization.

\begin{table}[h]
  \caption{
{ From-scratch augmentation ablation, $D=768$,} $R=4$, NFE 12.}
  \label{tab:appendix-augmentation-scratch}
  \vspace{3mm}
  \centering
  \small
  \begin{tabular}{lccc}
    \toprule
    Augmentation & Best Acc@1 (epoch) & Final Acc@1 & Runtime (h) \\
    \midrule
    Strong   & \textbf{74.22 (488)} & \textbf{74.22} & 47.77 \\
    Moderate & 73.09 (487) & 72.72 & 34.08 \\
    Weak     & 62.70 (452) & 62.27 & 33.26 \\
    \bottomrule
  \end{tabular}
\end{table}

\begin{table}[t]
  \caption{ImageNet$\to$CIFAR-100
{ transfer, $D=768$ backbone} from an ImageNet checkpoint, across all
  three architecture settings.}
  \label{tab:appendix-transfer}
  \vspace{3mm}
  \centering
  \small
  \begin{tabular}{lccc}
    \toprule
    Setting & Protocol / Augmentation & & Best Acc@1 \\
    \midrule
    Continuous & Linear probe (frozen) & Hard     & 66.26 (43) \\
               & Linear probe (frozen) & Weak     & \textbf{70.01 (45)} \\
               & Full fine-tune        & Hard     & 80.25 (91) \\
               & Full fine-tune        & Moderate & \textbf{83.84 (83)} \\
    \midrule
    Discrete   & Linear probe (frozen) & Hard     & 66.77 (48) \\
               & Linear probe (frozen) & Weak     & \textbf{69.05 (43)} \\
               & Full fine-tune        & Hard     & 86.71 (92) \\
               & Full fine-tune        & Moderate & \textbf{87.09 (95)} \\
    \midrule
    Looped     & Linear probe (frozen) & Hard     & 68.71 (48) \\
               & Linear probe (frozen) & Weak     & \textbf{70.96 (47)} \\
               & Full fine-tune        & Hard     & 82.98 (97) \\
               & Full fine-tune        & Moderate & \textbf{84.32 (99)} \\
    \bottomrule
  \end{tabular}
\end{table}

\begin{figure}[t]
  \centering
  \includegraphics[width=\linewidth]{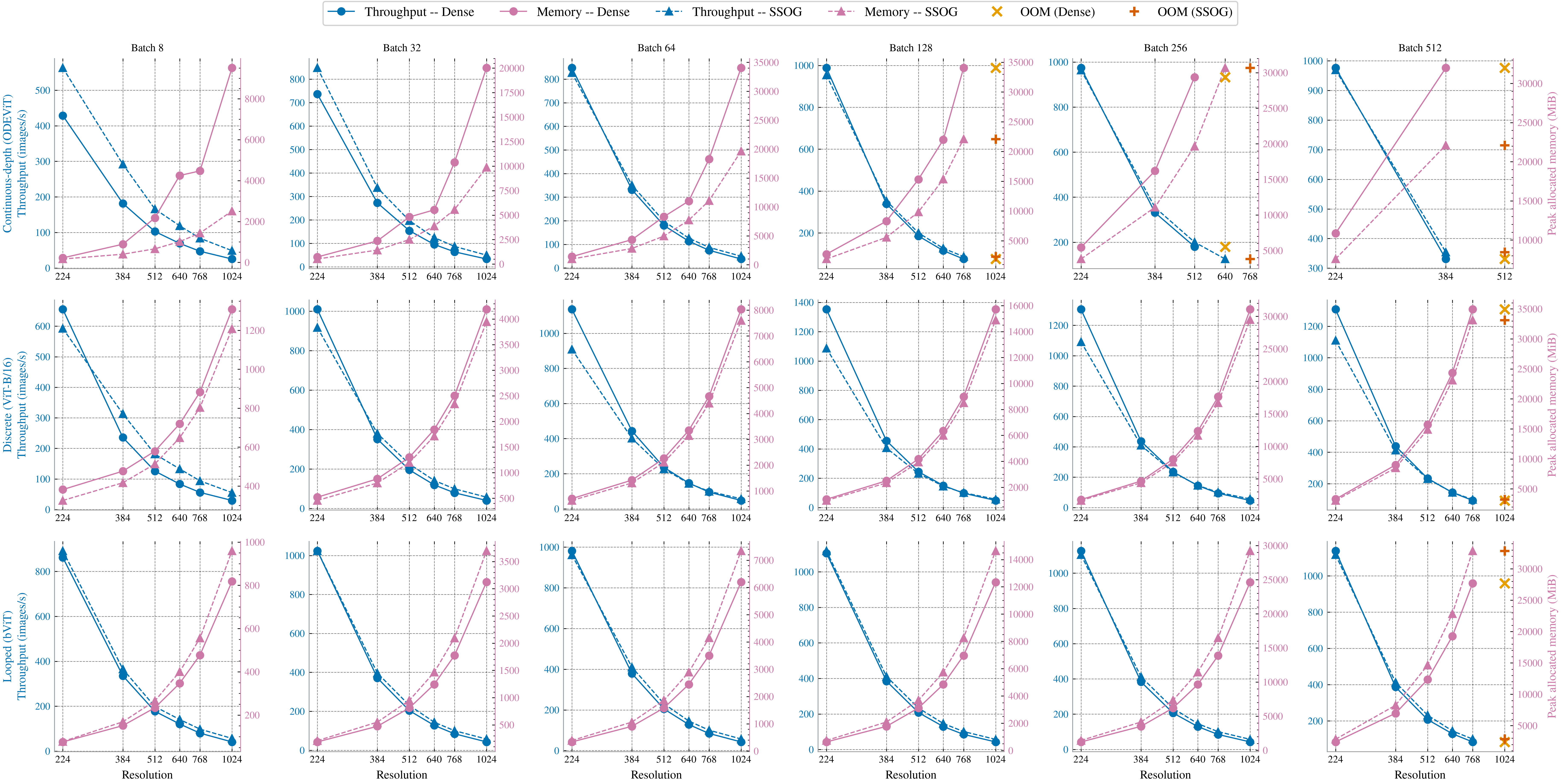}
  \caption{Throughput and peak memory versus input resolution, dense
  attention versus SSoG, across all three architecture settings (rows:
  continuous-depth ODE-ViT, discrete ViT-B/16, looped bViT) and six batch
  sizes (columns: 8, 32, 64, 128, 256, 512), at the paper's flagship width
  ($D=768$, $H=12$; NFE~12 for the continuous-depth setting, depth~12 for
  the discrete and looped settings). Each panel shares a log-scaled
  resolution axis (224-1024) with two y-axes: throughput in images/s
  (blue, left) and peak allocated memory in MiB (magenta, right). Solid
  circles/lines denote dense attention; dashed triangles denote SSoG.
  An $\times$ (dense) or $+$ (SSoG) marks the first resolution at which
  that configuration ran out of GPU memory; its vertical position is the
  last successfully measured value at that batch size, not a measured
  point at the marked resolution, and no further resolutions are plotted
  for that curve beyond it. All measurements are FP16 autocast (model
  parameters and inputs in FP32) on a single NVIDIA A40, randomly
  initialized weights, a single run per configuration (no repeated-seed
  or repeated-measurement uncertainty), 10 warmup and 50 measured
  iterations per point.}
  \label{fig:efficiency-dual-axis-grid-all-batches}
\end{figure}

\clearpage
\subsection{Lipschitz continuity of SSoG}
\label{sec:appendix-lipschitz}

In the continuous-depth setting of~\eqref{eq:method-ode}, the token matrix $X(t)$ evolves under the vector field $F(\cdot,\cdot|\theta)$. By the Picard--Lindel\"of theorem~\cite{coddington1955theory}, if $F$ is continuous in $t$ and Lipschitz continuous in $X$ with a constant independent of $t$ and of $X$, i.e.\ globally Lipschitz, then~\eqref{eq:method-ode} has a unique solution on the whole interval $[0,1]$ for every initial condition $X_0$. If $F$ is only Lipschitz on bounded sets, the same theorem guarantees existence and uniqueness only on a possibly shorter interval, which extends to $[0,1]$ provided the trajectory remains bounded.

The vector field $F(\cdot,\cdot|\theta)$ is built from the SSoG spatial operator, the value and output projections $W_V^h$ and $W_O$, the feed-forward block and layer normalization. Since sums and compositions of globally Lipschitz maps are globally Lipschitz, it suffices to bound each component. Linear maps are globally Lipschitz with constant equal to their spectral norm; the feed-forward block is globally Lipschitz with constant at most the product of the spectral norms of its weight matrices and the Lipschitz constant of its activation; and layer normalization with a stabilizing constant $\varepsilon_{\mathrm{LN}}>0$ is globally Lipschitz (see the discussion following Proposition~\ref{thm:fixed}). The only non-trivial component is therefore the SSoG spatial operator. We show that fixed SSoG is globally Lipschitz (Proposition~\ref{thm:fixed}), so that ODE-SSoG with fixed kernels has a unique solution on $[0,1]$, whereas content-steered SSoG is only Lipschitz on bounded sets (Proposition~\ref{thm:steered}), so that for this variant only the local guarantee applies.

{
\paragraph{Notation.}
We recall the notation of~\S\ref{sec:method}. For head $h$, the value projection is $V^h = XW_V^h$ (\eqref{eq:method-values}).
For each atom $r$ and axis $a\in\{x,y\}$, with $\delta_a := q_a - s_a$, let $g_{r,a}(\delta_a) := \exp\!\big(\log\Gker^{a}(\delta_a)/\tau\big)$ denote the temperature-scaled Gaussian kernel of atom $r$, where $\log\Gker^{a}$ is given by~\eqref{eq:method-gaussian-log-kernel} and $\tau$ is the temperature of~\eqref{eq:method-axis-kernel}. The normalized axis kernel $A_{r,a}(\delta_a)$ coincides with $A_a(\delta_a)$ from~\eqref{eq:method-axis-kernel} evaluated for atom $r$. The mixture weights $(\lambda_r)_r$ lie in the probability simplex (\eqref{eq:method-constraints}), and the per-block class-token aggregate is $v_{\mathrm{cls}} = \tfrac{1}{N-1}\sum_{p=1}^{N-1} v_p$ (\eqref{eq:method-cls-update}). All norms on token matrices are Frobenius; norms on linear maps are spectral.
}

We drop the batch index, since every operation is separable across the batch.

\begin{lemma}[Kernel norm]
\label{lem:axisnorm}
{ For each atom $r$ and each axis $a\in\{x,y\}$, the axis kernel of~\eqref{eq:method-axis-kernel} satisfies}
{ $\|A_{r,a}\|_2\le\sqrt{\kappa_{r,a}}$}, where
\begin{equation}
    { \kappa_{r,a}
    =
    \frac{\max_{s_a}\sum_{q_a} g_{r,a}(\delta_a)}
         {\min_{q_a}\sum_{s_a} g_{r,a}(\delta_a)},
    \quad \delta_a := q_a-s_a}
    \;\ge\;1
    \label{eq:kappa}
\end{equation}
and %
{ $g_{r,a}(\delta_a)$} is
{ the temperature-scaled Gaussian kernel $g_{r,a}(\delta_a)=\exp\!\big(\log\Gker^{a}(\delta_a)/\tau\big)$, with $\log\Gker^{a}$ as in~\eqref{eq:method-gaussian-log-kernel} evaluated for atom $r$ and $\tau$ the temperature of~\eqref{eq:method-axis-kernel}.} %
\end{lemma}

{
\begin{proof}
Since all entries of $A_{r,a}$ are positive, $\|A_{r,a}\|_\infty$ and $\|A_{r,a}\|_1$ are its maximum row sum and maximum column sum, respectively. We bound each in turn.

By~\eqref{eq:method-axis-kernel}, each row of $A_{r,a}$ is a softmax over the source coordinate $s_a$, so $\sum_{s_a}A_{r,a}(q_a,s_a)=1$ for every $q_a$. Hence $\|A_{r,a}\|_\infty=1$.

For a fixed column $s_a$, each entry $A_{r,a}(q_a,s_a)$ equals $g_{r,a}(q_a-s_a)$ divided by the row sum $\sum_{s_a'}g_{r,a}(q_a-s_a')$. Bounding every such denominator from below by its minimum over $q_a$ gives
\begin{equation*}
    \sum_{q_a} A_{r,a}(q_a,s_a)
    \;=\; \sum_{q_a} \frac{g_{r,a}(q_a-s_a)}{\displaystyle\sum_{s_a'} g_{r,a}(q_a-s_a')}
    \;\le\; \frac{\displaystyle\sum_{q_a} g_{r,a}(q_a-s_a)}{\displaystyle\min_{q_a}\sum_{s_a'}g_{r,a}(q_a-s_a')}.
\end{equation*}
The denominator on the right no longer depends on $s_a$, so taking the maximum over $s_a$ on both sides yields $\|A_{r,a}\|_1\le\kappa_{r,a}$, with $\kappa_{r,a}$ as in~\eqref{eq:kappa}.

Finally, every real matrix $M$ satisfies $\|M\|_2\le\sqrt{\|M\|_1\|M\|_\infty}$, since $\|M\|_2^2=\rho(M^\top M)\le\|M^\top M\|_\infty\le\|M^\top\|_\infty\|M\|_\infty=\|M\|_1\|M\|_\infty$. Applying this to $A_{r,a}$ with the two bounds above gives $\|A_{r,a}\|_2\le\sqrt{\kappa_{r,a}}$.
\end{proof}
}

{
\paragraph{Remark.}
The constant $\kappa_{r,a}$ is finite for every choice of atom parameters, which is all that Proposition~\ref{thm:fixed} requires; with truncated boundaries, however, it does not approach one as the grid grows. A query $q_a$ at the edge of the axis only retains the part of $g_{r,a}(q_a,\cdot)$ that falls inside the grid, whereas an interior column collects the full kernel mass. Consequently $\kappa_{r,a}\approx 1$ only when the effective width $\sqrt{\tau}\,\sigma_{r,a}$ is small compared with the patch spacing and the offset $\mu_{r,a}$ is close to zero; for a centred kernel of larger width $\kappa_{r,a}$ is close to $2$, and it grows rapidly with $|\mu_{r,a}|$, since edge rows then retain only a tail of the kernel. The bound $\sqrt{\kappa_{r,a}}$ is therefore conservative.
}

\begin{proposition}[Fixed SSoG is globally Lipschitz]
\label{thm:fixed}
{
Let $X\in\mathbb{R}^{N\times D}$ consist of one class token and $N-1=\gy\gx$ patch tokens arranged on an $\gy\times\gx$ grid. For each head $h\in\{1,\dots,H\}$, assume that the atom parameters $\mu_{r,a}$, $\sigma_{r,a}$, $\ell_r$ and the temperature $\tau$ are fixed, i.e.\ independent of $X$. Let $W_V^h\in\mathbb{R}^{D\times D^h}$ be the value projections of~\eqref{eq:method-values}, $W_V=W_V^1\mathbin{\|}\cdots\mathbin{\|}W_V^H\in\mathbb{R}^{D\times D}$ their concatenation, and $W_O\in\mathbb{R}^{D\times D}$ the output projection. Let
\begin{equation}
\mathcal{A}^h=\sum_{r=1}^{R}{ \lambda^h_r}\,\big(A^h_{r,y}\otimes A^h_{r,x}\big)\in\mathbb{R}^{(N-1)\times(N-1)}
    \label{eq:effective-kernel}
\end{equation}
be the effective spatial kernel of head $h$ on the patch tokens. Then fixed SSoG is a linear map on $\mathbb{R}^{N\times D}$ and, with the Frobenius norm on token matrices,
\begin{equation}
    \operatorname{Lip}(\mathrm{SSoG})
    \;\le\;
    \|W_V\|_2\,\|W_O\|_2\,K_{\mathcal{A}},
    \label{eq:lip-fixed}
\end{equation}
where
\begin{equation}
    K_{\mathcal{A}}
    :=
    \max_{h}\sqrt{\tfrac{1}{N-1}+\|\mathcal{A}^h\|_2^{2}}
    \;\le\;
    \max_{h}\sqrt{\tfrac{1}{N-1}+\max_{r}\kappa^h_{r,y}\,\kappa^h_{r,x}},
    \label{eq:K-fixed}
\end{equation}
and $\kappa^h_{r,a}$ is the constant of Lemma~\ref{lem:axisnorm} for atom $r$ of head $h$. The bound depends only on the parameters and not on the input.}
\end{proposition}

{
\begin{proof}
Fix a head $h$, order the tokens with the class token last, as in~\S\ref{sec:method}, and write $V^h=XW_V^h=\begin{pmatrix}V^h_{\mathrm{P}}\\ v^h_{\mathrm{cls}}\end{pmatrix}$, where $V^h_{\mathrm{P}}\in\mathbb{R}^{(N-1)\times D^h}$ collects the patch rows. Vectorizing the grid in row-major order and using $\operatorname{vec}(AUB^\top)=(A\otimes B)\operatorname{vec}(U)$, \eqref{eq:method-fixed-separable} gives the filtered patch rows $\mathcal{A}^hV^h_{\mathrm{P}}$, while by~\eqref{eq:method-cls-update} the class row is $\tfrac{1}{N-1}\mathbf{1}^\top V^h_{\mathrm{P}}$. Hence the output of head $h$ is $M^hXW_V^h$, with
\begin{equation*}
    M^h=\begin{pmatrix}\mathcal{A}^h & 0\\ \tfrac{1}{N-1}\mathbf{1}^\top & 0\end{pmatrix}\in\mathbb{R}^{N\times N}.
\end{equation*}

In the fixed variant $M^h$ does not depend on $X$. Denoting by
\begin{equation*}
    Z^h=M^hXW_V^h\in\mathbb{R}^{N\times D^h}
\end{equation*}
the output of head $h$, the SSoG output is obtained by concatenating the heads and applying the output projection,
\begin{equation*}
    \mathrm{SSoG}(X)=\big(Z^1\mathbin{\|}\cdots\mathbin{\|}Z^H\big)W_O ,
\end{equation*}
which is linear in $X$; its Lipschitz constant is therefore its operator norm. Since the heads occupy disjoint column blocks,
\begin{equation*}
\begin{split}
        \|\mathrm{SSoG}(X)\|_F
    \le&\|W_O\|_2\Big(\sum_h\|Z^h\|_F^2\Big)^{1/2}
    \le\|W_O\|_2\max_h\|M^h\|_2\,\|XW_V\|_F \\
    \le&\|W_V\|_2\|W_O\|_2\max_h\|M^h\|_2\,\|X\|_F .
\end{split}
\end{equation*}
Since the last column of $M^h$ vanishes, $\|M^h\|_2$ equals the spectral norm of its first $N-1$ columns; since, moreover, the squared spectral norm of a matrix equals the spectral norm of its Gram matrix,
\begin{equation*}
    \|M^h\|_2^2
    =\left\|\begin{pmatrix}\mathcal{A}^h\\ \tfrac{1}{N-1}\mathbf{1}^\top\end{pmatrix}\right\|_2^2
    =\Big\|(\mathcal{A}^h)^\top\mathcal{A}^h+\tfrac{1}{(N-1)^2}\mathbf{1}\mathbf{1}^\top\Big\|_2
    \le\|\mathcal{A}^h\|_2^2+\tfrac{1}{N-1},
\end{equation*}
where the inequality follows from the triangle inequality and $\|\mathbf{1}\mathbf{1}^\top\|_2=N-1$,
so that $\max_h\|M^h\|_2\le K_{\mathcal{A}}$, which gives~\eqref{eq:lip-fixed}.
For the inequality in~\eqref{eq:K-fixed}, $\|A\otimes B\|_2=\|A\|_2\|B\|_2$ and
{ $(\lambda^h_r)_r$} lies in the simplex by~\eqref{eq:method-constraints}, so~\eqref{eq:effective-kernel} is a convex combination and, by Lemma~\ref{lem:axisnorm}, $\|\mathcal{A}^h\|_2\le\max_r\|A^h_{r,y}\|_2\|A^h_{r,x}\|_2\le\max_r\sqrt{\kappa^h_{r,y}\kappa^h_{r,x}}$.
\end{proof}}

{ Proposition~\ref{thm:fixed} has two immediate consequences. For the discrete and looped models, the residual block of~\eqref{eq:method-discrete} is globally Lipschitz with constant at most $1+\|W_V\|_2\|W_O\|_2K_{\mathcal{A}}$. Since $K_{\mathcal{A}}$ is finite for any choice of parameters, the only factors not determined by the kernel are the projection norms, which spectral normalization bounds explicitly and weight decay keeps moderate in practice. For ODE-SSoG, the vector field $F(\cdot,\cdot|\theta)$ of~\eqref{eq:method-ode} composes fixed SSoG with the feed-forward block and layer normalization. The feed-forward block is globally Lipschitz, with constant at most the product of the spectral norms of its weight matrices and the Lipschitz constant of its activation. Layer normalization with gain $\gamma$ and stabilizing constant $\varepsilon_{\mathrm{LN}}>0$ added to the variance has Jacobian of spectral norm at most $\max_i|\gamma_i|/\sqrt{\varepsilon_{\mathrm{LN}}}$, so it is globally Lipschitz as well. Hence $F(\cdot,\cdot|\theta)$ is continuous in $t$ and globally Lipschitz in $X$ with a constant independent of $t$, and the Picard--Lindel\"of theorem yields a unique solution of~\eqref{eq:method-ode} on $[0,1]$.}

{ We now turn to content-steered SSoG. Steering acts on each head independently: the predictors of~\eqref{eq:method-steering} modify only the atom parameters of their own head, and different heads interact only through the output projection $W_O$, which is handled exactly as in Proposition~\ref{thm:fixed}. We therefore first analyse a single head. In Lemma~\ref{lem:steering} and its proof we fix one head and, to lighten notation, omit the superscript $h$ from all of its quantities; the superscript is restored in Proposition~\ref{thm:steered}, where all heads are combined.}

\paragraph{Intuition and notation for Lemma~\ref{lem:steering}.}
Steering can only move each atom within a bounded range. Since $|\tanh|<1$, \eqref{eq:method-steering} shifts the center $\mu_{r,a}$ by at most $\alpha_\mu M$ and multiplies the spread $\sigma_{r,a}$ by a factor between $e^{-\alpha_\sigma}$ and $e^{\alpha_\sigma}$. Together with the floor $\sigma_{r,a}\ge\sigma_{\min}$, this confines every steered spread to an interval $[\sigma_\star,\sigma^\star]$, and every distance between a grid offset $\delta_a\in\{-(n_a-1),\dots,n_a-1\}$ and a steered center to at most $\Delta$, where%
\begin{equation}
    \sigma_\star=\sigma_{\min}e^{-\alpha_\sigma},\qquad
    \sigma^\star=\max_{r,a}\sigma_{r,a}e^{\alpha_\sigma},\qquad
    \Delta=\max_{r,a}\big(n_a-1+|\mu_{r,a}|\big)+\alpha_\mu M.
    \label{eq:steer-range}
\end{equation}
Thus $\sigma_\star$ is the narrowest spread an atom can reach, $\sigma^\star$ the widest, and $\Delta$ the largest distance between a steered center and a grid offset. The first and the last bound how fast the logits of the axis softmax, $\log g_{r,a}(\delta_a)$, react to the atom parameters: on the logarithmic scale a Gaussian is a parabola, whose slope grows as the spread shrinks and as the offset moves away from the center, so the worst case is a spread $\sigma_\star$ at distance $\Delta$. The kernel value itself is close to zero far from the center, and the softmax damps changes in small entries, so bounding the kernel through its logits is conservative. The widest spread $\sigma^\star$ matters for a different reason: steering rescales the spread multiplicatively, so the same change in the predictor output moves a wide atom more than a narrow one. Lemma~\ref{lem:steering} turns these bounds into a Lipschitz constant, namely how fast the atom parameters follow the query token times how fast the kernel follows its parameters.

\begin{lemma}[Steered kernel]
\label{lem:steering}
{
Consider a single head, and let its atom parameters be steered by the query token $x_q$ as in~\eqref{eq:method-steering}, with steering radius $M$, predictor weights $W_\mu,W_\sigma,W_\lambda$ and scales $\alpha_\mu,\alpha_\sigma,\alpha_\lambda>0$ as in~\eqref{eq:method-steering-scales}, and with $\sigma_{r,a}\ge\sigma_{\min}$ and $\tau\ge\tau_0$ as in~\eqref{eq:method-constraints}.
Denote by $A_{r,a}(x_q)\in\mathbb{R}^{n_a}$ the row of $A_{r,a}$ associated with the query position $q$, and by $\mathcal{A}(x_q)=\sum_r\lambda_r(x_q)\,A_{r,y}(x_q)\otimes A_{r,x}(x_q)$ the corresponding row of the effective kernel. Let $\sigma_\star$, $\sigma^\star$ and $\Delta$ be as in~\eqref{eq:steer-range}. Then:
\begin{enumerate}
\renewcommand{\labelenumi}{(\roman{enumi})}
\item for every $x_q$, $r$ and $a$,
\begin{equation}\big(\mu_{r,a}(x_q),\sigma_{r,a}(x_q)\big)\in\Theta_{r,a}:=[\mu_{r,a}-\alpha_\mu M,\,\mu_{r,a}+\alpha_\mu M]\times[\sigma_\star,\sigma^\star];
\end{equation}
\item the row $\mathcal{A}(x_q)$ of the effective kernel is Lipschitz in $x_q$ with respect to the Euclidean norm, with constant
\begin{equation}
    L_{\mathcal{A}}=\tfrac12\sqrt{R/2}\;\alpha_\lambda\|W_\lambda\|_2+\max_r\big(L_{r,y}+L_{r,x}\big),
    \label{eq:LA}
\end{equation}
where
\begin{equation}
    L_{r,a}=\frac{\sqrt{n_a}}{2\tau_0}\left(\frac{\Delta}{\sigma_\star^{2}}\,\alpha_\mu M\|W_\mu\|_2+\Big(\frac{1}{\sigma_\star}+\frac{\Delta^{2}}{\sigma_\star^{3}}\Big)\sigma^\star\alpha_\sigma\|W_\sigma\|_2\right).
    \label{eq:Lra}
\end{equation}
\end{enumerate}}
\end{lemma}

{
\begin{proof}
(i) Since $|\tanh|\le1$ and $\sigma_{r,a}\ge\sigma_{\min}$, $|\mu_{r,a}(x_q)-\mu_{r,a}|\le\alpha_\mu M$ and $\sigma_\star\le\sigma_{r,a}(x_q)\le\sigma^\star$.

(ii) Fix two tokens $x_q$ and $x'_q$. For the center, write $u=[W_\mu x_q+b_\mu]_{r,a}$ and $u'=[W_\mu x'_q+b_\mu]_{r,a}$ for the pre-activations of its predictor, so that $|u-u'|\le\|W_\mu\|_2\,\|x_q-x'_q\|_2$. Since $\tanh$ is $1$-Lipschitz,%
\begin{equation*}
    |\mu_{r,a}(x_q)-\mu_{r,a}(x'_q)|
    =\alpha_\mu M\,|\tanh u-\tanh u'|
    \le\alpha_\mu M\,\|W_\mu\|_2\,\|x_q-x'_q\|_2 .
\end{equation*}
For the spread, write likewise $v=[W_\sigma x_q+b_\sigma]_{r,a}$ and $v'=[W_\sigma x'_q+b_\sigma]_{r,a}$, and regard the steered spread as the function $v\mapsto\sigma_{r,a}\exp(\alpha_\sigma\tanh v)$. Its derivative satisfies%
\begin{equation*}
    0\le\frac{\partial\sigma_{r,a}}{\partial v}
    =\sigma_{r,a}\,e^{\alpha_\sigma\tanh v}\,\alpha_\sigma\tanh'(v)
    \le\sigma^\star\alpha_\sigma ,
\end{equation*}
because the steered spread never exceeds $\sigma^\star$ and $0<\tanh'\le1$. By the mean value theorem,%
\begin{equation*}
    |\sigma_{r,a}(x_q)-\sigma_{r,a}(x'_q)|
    \le\sigma^\star\alpha_\sigma\,|v-v'|
    \le\sigma^\star\alpha_\sigma\,\|W_\sigma\|_2\,\|x_q-x'_q\|_2 .
\end{equation*}

The axis row $A_{r,a}(x_q)$ is the softmax of $z\in\mathbb{R}^{n_a}$, $z_s=\log g_{r,a}(q_a-s)$, where $\log g_{r,a}=\tau^{-1}\log\Gker^{a}$, with $\log\Gker^{a}$ as in~\eqref{eq:method-gaussian-log-kernel}, is evaluated at $\mu_{r,a}(x_q)$ and $\sigma_{r,a}(x_q)$; let $z'$ be the same vector at $x'_q$. On $\Theta_{r,a}$, every grid offset satisfies $|\delta_a-\mu|\le\Delta$, so $\tau\ge\tau_0$ gives
\begin{equation}
    \left|\frac{\partial\log g_{r,a}}{\partial\mu}\right|\le\frac{\Delta}{\tau_0\,\sigma_\star^{2}},
    \qquad
    \left|\frac{\partial\log g_{r,a}}{\partial\sigma}\right|\le\frac{1}{\tau_0}\left(\frac{1}{\sigma_\star}+\frac{\Delta^{2}}{\sigma_\star^{3}}\right).
    \label{eq:kernel-derivs}
\end{equation}
Since $\Theta_{r,a}$ is convex, the segment joining $\big(\mu_{r,a}(x_q),\sigma_{r,a}(x_q)\big)$ and $\big(\mu_{r,a}(x'_q),\sigma_{r,a}(x'_q)\big)$ stays in $\Theta_{r,a}$, where the bounds~\eqref{eq:kernel-derivs} hold. The mean value theorem along this segment, together with the two estimates above, gives for every source coordinate $s$%
\begin{equation}
\begin{split}
    |z_s-z'_s|
    \le &\frac{\Delta}{\tau_0\sigma_\star^{2}}\,\big|\mu_{r,a}(x_q)-\mu_{r,a}(x'_q)\big|
    \\
    + &\frac{1}{\tau_0}\Big(\frac{1}{\sigma_\star}+\frac{\Delta^{2}}{\sigma_\star^{3}}\Big)\big|\sigma_{r,a}(x_q)-\sigma_{r,a}(x'_q)\big|
    \le\frac{2}{\sqrt{n_a}}\,L_{r,a}\,\|x_q-x'_q\|_2 ,
    \end{split}
\end{equation}
where the last step is the definition~\eqref{eq:Lra} of $L_{r,a}$. It remains to pass from the logits $z$ to the normalized row. The softmax is $\tfrac12$-Lipschitz in the Euclidean norm, because its Jacobian $\operatorname{diag}(p)-pp^\top$ is positive semidefinite with spectral norm at most $\tfrac12$. Hence%
\begin{equation}
    \|A_{r,a}(x_q)-A_{r,a}(x'_q)\|_2
    \le\tfrac12\|z-z'\|_2
    \le\tfrac12\sqrt{n_a}\,\max_s|z_s-z'_s|
    \le L_{r,a}\,\|x_q-x'_q\|_2 .
\end{equation}

For the effective kernel, write $c_r=A_{r,y}(x_q)\otimes A_{r,x}(x_q)$, $c_r'$ for its value at $x'_q$, and $d=\lambda(x_q)-\lambda(x'_q)$, so that
\begin{equation}
    \mathcal{A}(x_q)-\mathcal{A}(x'_q)=\sum_rd_r\,c_r+\sum_r\lambda_r(x'_q)\,(c_r-c_r').
\end{equation}
We bound the two sums separately.

{\em Second sum.} All rows are probability vectors, so their Euclidean norm is at most one. For any vectors $a,a',b,b'$,%
\begin{equation}
    \|a\otimes b-a'\otimes b'\|_2\le\|a-a'\|_2\,\|b\|_2+\|a'\|_2\,\|b-b'\|_2 ,
\end{equation}
and applying it with $a=A_{r,y}(x_q)$, $b=A_{r,x}(x_q)$ and their counterparts at $x'_q$ gives%
\begin{equation}
    \|c_r-c_r'\|_2\le\big(L_{r,y}+L_{r,x}\big)\,\|x_q-x'_q\|_2 .
\end{equation}
Since the weights $\lambda_r(x'_q)$ form a convex combination, the second sum is at most $\max_r(L_{r,y}+L_{r,x})\,\|x_q-x'_q\|_2$.

{\em First sum.} Its coefficients add up to zero, $\sum_rd_r=0$, so splitting $d$ into its positive and negative parts gives $\sum_rd_rc_r=\tfrac12\|d\|_1\,(P-Q)$ for two probability vectors $P$ and $Q$. Since $\|P-Q\|_2^2\le\|P-Q\|_1\|P-Q\|_\infty\le2$,%
\begin{equation}
    \Big\|\sum_rd_rc_r\Big\|_2\le\frac{\|d\|_1}{\sqrt2}\le\sqrt{R/2}\,\|d\|_2 .
\end{equation}
Finally, the mixture weights are a softmax of $\ell_r+\alpha_\lambda\tanh(\cdot)$; since the softmax is $\tfrac12$-Lipschitz and $\tanh$ is $1$-Lipschitz,%
\begin{equation}
    \|d\|_2\le\tfrac12\,\alpha_\lambda\,\|W_\lambda\|_2\,\|x_q-x'_q\|_2 .
\end{equation}
Adding the bounds on the two sums gives~\eqref{eq:LA}.
\end{proof}}

{ The constants $L_{r,a}$ and $L_{\mathcal{A}}$ of Lemma~\ref{lem:steering} are finite because of three design choices of~\S\ref{sec:method} and the finiteness of the grid: the floor $\sigma_{\min}$ keeps the spread away from zero, the steering radius $M$ bounds how far the center moves, the offset $\tau_0$ keeps the temperature away from zero, and $\delta_a$ ranges over a finite set. Removing any one of them makes the corresponding bound in~\eqref{eq:kernel-derivs}, and hence $L_{r,a}$, diverge.}

\begin{proposition}[Steered SSoG is Lipschitz on bounded sets]
\label{thm:steered}
{
Let $\rho>0$, let $X$, $W_V^h$, $W_V$ and $W_O$ be as in Proposition~\ref{thm:fixed}, and let the atom parameters of every head $h$ be steered as in~\eqref{eq:method-steering}, under the assumptions of Lemma~\ref{lem:steering}, so that the effective kernel $\mathcal{A}^h(X)\in\mathbb{R}^{(N-1)\times(N-1)}$ has rows $\mathcal{A}^h(x_q)$ as in that lemma. For each head $h$, atom $r$ and axis $a$, let $\Theta^h_{r,a}$ be the set $\Theta_{r,a}$ of Lemma~\ref{lem:steering} for head $h$, and let
\begin{equation}
    \bar\kappa^h_{r,a}=\frac{\max_{s_a}\sum_{q_a}\sup_{\theta\in\Theta^h_{r,a}}g_\theta(q_a-s_a)}{\min_{q_a}\inf_{\theta\in\Theta^h_{r,a}}\sum_{s_a}g_\theta(q_a-s_a)},
    \label{eq:kappa-bar}
\end{equation}
where $g_\theta$ denotes the kernel $g_{r,a}$ of head $h$ evaluated at $\theta=(\mu,\sigma)$. Then, for all $X,X'$ with $\|X\|_F,\|X'\|_F\le\rho$,
\begin{equation}
    \|\mathrm{SSoG}(X)-\mathrm{SSoG}(X')\|_F
    \le
    \|W_V\|_2\|W_O\|_2\Big(\bar K_{\mathcal{A}}+\max_h L^h_{\mathcal{A}}\,\rho\Big)\|X-X'\|_F,
    \label{eq:lip-steered}
\end{equation}
where
\begin{equation}
    \bar K_{\mathcal{A}}
    :=
    \max_h\sqrt{\tfrac{1}{N-1}+\textstyle\sum_r\bar\kappa^h_{r,y}\bar\kappa^h_{r,x}}
    \label{eq:K-steered}
\end{equation}
and $L^h_{\mathcal{A}}$ is the constant~\eqref{eq:LA} for head $h$. In particular, the Lipschitz bound on the ball of radius $\rho$ grows linearly in $\rho$.}
\end{proposition}

{
\begin{proof}
With the class token last, as in the proof of Proposition~\ref{thm:fixed}, let
$M^h(X)=\begin{pmatrix}\mathcal{A}^h(X)&0\\ \tfrac{1}{N-1}\mathbf{1}^\top&0\end{pmatrix}$. Denoting by
\begin{equation*}
    Z^h(X)=M^h(X)XW_V^h\in\mathbb{R}^{N\times D^h}
\end{equation*}
the output of head $h$, the SSoG output is obtained by concatenating the heads and applying the output projection,
\begin{equation*}
    \mathrm{SSoG}(X)=\big(Z^1(X)\mathbin{\|}\cdots\mathbin{\|}Z^H(X)\big)W_O .
\end{equation*}
Since the class row of $M^h$ does not depend on $X$,
\begin{equation*}
    Z^h(X)-Z^h(X')
    =M^h(X)(X-X')W_V^h+\begin{pmatrix}\big(\mathcal{A}^h(X)-\mathcal{A}^h(X')\big)X'_{\mathrm{P}}W_V^h\\ 0\end{pmatrix},
\end{equation*}
where $X'_{\mathrm{P}}$ collects the patch rows of $X'$.
Arguing over the heads as in the proof of Proposition~\ref{thm:fixed} and using the triangle inequality, the first terms contribute at most: %
$$\|W_V\|_2\|W_O\|_2\max_h\|M^h(X)\|_2\,\|X-X'\|_F$$ %
and, since $\|X'_{\mathrm{P}}W_V\|_F\le\|W_V\|_2\|X'\|_F$, the second at most: %
$$\|W_V\|_2\|W_O\|_2\max_h\|\mathcal{A}^h(X)-\mathcal{A}^h(X')\|_2\,\|X'\|_F.$$

By Lemma~\ref{lem:steering}(i), every steered parameter lies in $\Theta^h_{r,a}$, a compact set on which $g_\theta$ is continuous and positive on the finite grid, so $\bar\kappa^h_{r,a}<\infty$. Every row of $\mathcal{A}^h(X)$ is a probability vector, so $\|\mathcal{A}^h(X)\|_\infty=1$.
Each entry of a row of $A^h_{r,a}$ is at most %
$$\sup_{\theta\in\Theta^h_{r,a}}g_\theta(q_a-s_a)\big/\min_{q_a}\inf_{\theta\in\Theta^h_{r,a}}\sum_{s_a}g_\theta(q_a-s_a)$$ %
and $\lambda^h_r\le1$ since $\lambda^h$ is a softmax, so the column of $\mathcal{A}^h(X)$ indexed by $s=(s_y,s_x)$ sums to at most $\sum_r\bar\kappa^h_{r,y}\bar\kappa^h_{r,x}$. Hence $$\|\mathcal{A}^h(X)\|_2\le(\|\mathcal{A}^h(X)\|_1\|\mathcal{A}^h(X)\|_\infty)^{1/2}\le\big(\sum_r\bar\kappa^h_{r,y}\bar\kappa^h_{r,x}\big)^{1/2}$$ and, exactly as in Proposition~\ref{thm:fixed}, $$\|M^h(X)\|_2\le\big(\tfrac{1}{N-1}+\sum_r\bar\kappa^h_{r,y}\bar\kappa^h_{r,x}\big)^{1/2},$$ %
so that $\max_h\|M^h(X)\|_2\le\bar K_{\mathcal{A}}$. By Lemma~\ref{lem:steering}(ii), %
$$\|\mathcal{A}^h(X)-\mathcal{A}^h(X')\|_2\le\|\mathcal{A}^h(X)-\mathcal{A}^h(X')\|_F=\big(\sum_q\|\mathcal{A}^h(x_q)-\mathcal{A}^h(x'_q)\|_2^2\big)^{1/2}\le L^h_{\mathcal{A}}\|X-X'\|_F.$$
Combining the two contributions,
\begin{equation*}
    \|\mathrm{SSoG}(X)-\mathrm{SSoG}(X')\|_F
    \le\|W_V\|_2\|W_O\|_2\Big(\bar K_{\mathcal{A}}+\max_hL^h_{\mathcal{A}}\,\|X'\|_F\Big)\|X-X'\|_F,
\end{equation*}
and $\|X'\|_F\le\rho$ gives~\eqref{eq:lip-steered}.
\end{proof}}

{
\paragraph{Remark.}
The dependence on $\rho$ in~\eqref{eq:lip-steered} cannot be removed in general. For non-degenerate steering, fix a query position $q$ and its token $x_q$, and scale the remaining patch tokens by a factor $c$: a perturbation of $x_q$ changes the kernel row $\mathcal{A}^h(x_q)$ at a rate that does not depend on $c$, while the values it multiplies are of size $O(c)$, so the directional derivative of steered SSoG grows linearly in $c$. When the steering does not act, for instance when $W_\mu=W_\sigma=W_\lambda=0$, the operator reduces to fixed SSoG and Proposition~\ref{thm:fixed} applies.}

{ The preceding Remark has the same structure as the obstruction that \citet{kim2021lipschitzconstantselfattention} identify for dot-product attention, namely a data-dependent mixing matrix multiplying a data-dependent value matrix, but not the same rate. Attention logits are inner products between two token features, so the Lipschitz constant of the mixing matrix itself grows with $\|X\|$ and the available upper bounds on a ball of radius $\rho$ grow as $\rho^{2}$; the steered kernel depends only on the query token and only through a bounded $\tanh$, so $L^h_{\mathcal{A}}$ does not depend on $X$ and the bound of Proposition~\ref{thm:steered} grows as $\rho$. Relative to dot-product attention, steering thus lowers the exponent of this bound from two to one, and the fixed variant removes the dependence on $\rho$ altogether. We therefore regard Proposition~\ref{thm:fixed} as a further reason to prefer the fixed operator, alongside the accuracy results of~\S\ref{sec:results-steering}.}

\end{document}

%% file: math_commands.tex
\usepackage{amsmath,amsfonts,bm}

\def\eqref#1{equation~\ref{#1}}

\def\1{\bm{1}}

\DeclareMathAlphabet{\mathsfit}{\encodingdefault}{\sfdefault}{m}{sl}
\SetMathAlphabet{\mathsfit}{bold}{\encodingdefault}{\sfdefault}{bx}{n}

